%% file: interpolants-preprint.tex
\documentclass[a4paper]{article}

\usepackage[a4paper,margin=0.9in]{geometry}
\usepackage[utf8]{inputenc} 
\usepackage[T1]{fontenc}    
\usepackage[hypertexnames=false,backref=page]{hyperref}
\usepackage{url}            
\usepackage{booktabs}       
\usepackage{amsfonts}       
\usepackage{nicefrac}       
\usepackage{microtype}      
\usepackage{xcolor}
\usepackage{algorithm}
\usepackage{algpseudocode}
\usepackage{amsmath, amsfonts, amssymb, bm, bbm}
\usepackage{subcaption}
\usepackage{wrapfig}
\usepackage{amsthm}
\usepackage{wrapfig}
\usepackage{thmtools, thm-restate}
\usepackage[textsize=tiny]{todonotes}
\usepackage[nameinlink,capitalize]{cleveref}
\usepackage[sort, round]{natbib}
\usepackage[backgroundcolor=orange!10]{mdframed}

\usepackage{localmacros}

\hypersetup{
    colorlinks,
    linkcolor={red!50!black},
    citecolor={blue!50!black},
    urlcolor={green!40!black}
}

\theoremstyle{plain}

\theoremstyle{definition}
\newtheorem{definition}{Definition}
\newtheorem{assumption}{Assumption}

\newtheorem{proposition}{Proposition}[section]
\newtheorem{lemma}{Lemma}[section]

\newtheorem*{proposition*}{Proposition}
\newtheorem*{lemma*}{Lemma}

\numberwithin{equation}{section}

\graphicspath{
    {./figures/}
    }

\author{%
Mario Massy$^{1}$ 
 \quad Sara Shamekh$^{2, 3}$   \\
\quad  Fabrizio Falasca $^{2, *}$
\quad  Matthieu Blanke$^{2,3,4, *}$ \\ \\
$^{1}$École polytechnique, Institut polytechnique de Paris \\
$^{2}$Courant Institute, New York University \\
$^{3}$LEAP NSF STC \\
$^{4}$LMD, ISPL, ENS, Université PSL\\
$^{*}$Equal advising
}

\title{Stable probabilistic emulation of
\\
partially observed dynamical systems}

\begin{document}

\maketitle
\begin{abstract}
    \noindent
    Predicting the long-term behavior of partially observed dynamical systems such as the coupled atmosphere-ocean system is central to societal decision-making.
    Motivated by the success of deep learning emulators in weather forecasting, data-driven climate models represent a promising approach.
    However, climate emulation poses distinct challenges: models must account for unresolved fast variables introduced by unavoidable coarse-graining, reproduce long-term statistics, and remain stable over arbitrarily long rollouts and under external perturbations. These requirements motivate stochastic data-driven emulators, but modern generative models lack stability guarantees. We introduce forced-dissipative stochastic interpolants, a generative framework for learning stochastic reduced dynamics. Its architecture uses structural constraints inspired by turbulent dynamics, which provide stability by design while retaining the flexibility of stochastic interpolants to represent complex probabilistic dynamics. Building on the theory of Markov-chain stability, we establish a Foster--Lyapunov drift condition and prove geometric ergodicity for our emulator, together with stability under steady external forcing. We evaluate the framework on the two-timescale Lorenz-96 system, a prototype of partially observed multiscale turbulence. The resulting emulator remains stable over long integrations and under finite-amplitude forcing, with accurate prediction of forced responses of both the mean and variance.
\end{abstract}

\section{Introduction} \label{sec:intro}
A central challenge across many scientific domains is to build reliable data-driven models of complex dynamical systems from partial observations.
A paradigmatic example is climate modeling, where coupled ocean--atmosphere dynamics evolve over a wide range of spatial and temporal scales \citep{MajdaBranicki,GhilLucarini}, and where numerical models resolving these scales are costly to run.
Motivated by the success of data-driven weather forecasting \citep{gencast,Lam2023,Bi2023,fourcastnet3}, neural climate emulators have recently been developed as autoregressive models that learn directly from data, and produce predictions at a fraction of the cost of the numerical models they emulate \citep{Kochkov2023,Watt-Meyer2023,Watt-Meyer2024,Lucie,OceanNet,Bire,Samudra,Camulator,lucie3D,SamudraACE}.
Since only partial, coarse-grained observations are available (\textit{e.g.}, measurements of temperature or large-scale winds at finite resolution), the aggregate effect of unresolved and unobserved small scales on the resolved variables must be taken into account, and stochastic or probabilistic modeling naturally arises \citep{Penland89,MajdaBook,NormalForms,majdaSIAM,KONDRASHOV201533,FabriCoarseGraining,GiorginiScore1,GiorginiScore2,Freitas,Brolly}.
Generative models such as diffusion models and stochastic interpolants then provide a flexible framework to learn such probabilistic emulators from data alone \citep{kohl2023acdm, gencast,pmlr-v235-chen24n,ProbabilisticSIPedram}.

Unlike weather forecasting, climate emulation does not aim at accurate short-term forecasts: useful emulators should reproduce the stationary statistics of the system produced by long autoregressive rollouts, and predict its responses to external perturbations.
Long-term stability of neural emulators is challenging because of their autoregressive nature: small errors can accumulate over long rollouts, eventually leading to finite-time blow-up or nonphysical behavior \citep{ainslie2026eigenanalysis}. Proposed solutions to this problem include corrective procedures \citep{Chris,QiLiu}, training or predicting over more than one temporal step \citep{Stachenfeld,Keisler}, Fourier-space regularizations and higher-order integration schemes \citep{chattopadhyay2023instabilities}.  In the context of stochastic interpolants, \citet{Nikolaj} recently proposed soft energy constraints to match the system's energy.
While the strategies above effectively improve the long-run behavior, they are not supported by a formal mathematical stability guarantee, especially for probabilistic models.
Furthermore, stability of long rollouts and an accurate representation of stationary statistics are not sufficient for a model to capture physically meaningful behavior. A reliable emulator must correctly capture the system's response to external perturbations, for key statistics such as the mean and variance,
which are shown to depend critically on the stochastic parameterization of unresolved processes \citep{majdaSIAM,FabriCoarseGraining,GiorginiScore2}.
Examining the emulator's response provides a stringent test of whether it has learned the dynamical mechanisms governing the system, and can reveal physical inconsistencies and spurious causal links \citep{FabriCoarseGraining,falasca2026physics}. It is particularly challenging for data-driven emulators, because perturbations inevitably drive the learning model out of its training distribution, where neural networks are most likely to go unstable. Although crucial in practice, the response to perturbations is rarely analyzed. In this work, we address these limitations by introducing a probabilistic emulator whose stability is guaranteed mathematically, including under external forcing, and is empirically accurate.

\paragraph{Contributions.}
Motivated by challenges of data-driven climate emulation, we formulate the problems of long-term stability and perturbation response of probabilistic autoregressive emulators.
We then introduce forced-dissipative stochastic interpolants, a generative architecture that uses structural properties of forced-dissipative turbulent dynamics as a blueprint for a hard architectural constraint, while retaining the flexibility of stochastic interpolants to learn probabilistic transitions from data. We prove that our generative process satisfies a Foster--Lyapunov drift condition, and show that it implies stability for the resulting emulator.
Our approach is illustrated on the two-scale Lorenz-96 system, a prototype of partially observed geophysical turbulence.
We show that our emulator remains stable where unconstrained stochastic interpolants and diffusion models become unstable, and it accurately predicts the system's stationary statistics and its response to large external forcings.

\section{Problem formulation for data-driven generative emulation} \label{sec:problem}

In this section, we state the mathematical problem of data-driven probabilistic emulation, relying on Markov chains, and introduce the stochastic interpolant framework we will build on.

\subsection{Probabilistic emulators}
\label{sec:probabilistic_emulators}

A discrete time series of consecutive observations $\mathcal{D}:= (y_1, \dots, y_N)$ in $\R^d$ is available, recorded from a physical system observed at a coarse level of description. In this context, the state $y_t$ does not determine $y_{t+1}$, as the unresolved scales act on the observed variables as an effective noise.  Assuming that these unresolved effects decorrelate fast over the observation interval, these partially observed states can then be modeled as realizations of a time-homogeneous Markov chain~$(Y_t)$, with transitions governed by a conditional probability density $p_\star$, as
\begin{equation}  \label{eq:true_markov_chain}
    Y_{t+1} \sim p_\star(\cdot \mid Y_t=y_t)
    \quad \text{for all} \; t \in \N.
\end{equation}
We assume throughout that the data-generating chain \eqref{eq:true_markov_chain} is stationary and that its invariant measure admits a probability density $\mu_\star$.

\paragraph{Probabilistic emulation.}
In this partially observed setting, a deterministic predictor $y_t \mapsto y_{t+1}$ is ill-suited, as it regresses toward the conditional mean, and its autoregressive iteration produces unphysically smooth trajectories.
Rather, probabilistic data-driven autoregressive emulators aim to learn a conditional generative model $p(\cdot \mid x)$ from $\mathcal{D}$, approximating the conditional data distribution~$p_\star(\cdot \mid x)$.
Once learned, forecasts are produced by applying the generative model autoregressively:
\begin{equation} \label{eq:emulated_trajectory}
    X_{t+1} \sim p(\cdot \mid X_t)
    \quad \text{for all} \; t \in \N,
\end{equation}
which defines a time-homogeneous Markov chain on $\R^d$. We denote its transition kernel by $P$, so that $P(x, \cdot)$ is the distribution of the next state given the current state $x$, and the distribution of the state $t$ steps later is denoted $P^t(x, \cdot)$.

Before introducing a specific architecture for the generative modeling process~\eqref{eq:emulated_trajectory}, we discuss the mathematical framework for evaluating probabilistic emulators.

\subsection{Stability and perturbations}
\label{sec:stability}

An emulator useful for climate modeling should reproduce the system's long-term statistics and its response to external forcings. Both objectives first require stable rollouts. The long-term statistics of the emulated chain \eqref{eq:emulated_trajectory} are described by an invariant probability measure $\mu$, toward which the chain equilibrates. For a learned kernel, the existence of such a measure is not guaranteed. We now define stability and perturbations formally.

\begin{definition}[Stability]\label{definition:stability}
    A Markov chain with transition kernel $Q$ is {stable} if it admits a unique invariant probability measure $\mu$ and converges to it in total variation from every initial condition:
    \begin{equation} \label{eq:stability_definition}
        \big\Vert Q^t(x, \cdot) - \mu \big\Vert_{\mathrm{TV}}
        \xrightarrow[t \to + \infty]{} 0
        \quad \text{for all} \; x \in \R^d,
    \end{equation}
    where $\Vert \cdot \Vert_{\mathrm{TV}}$
    denotes the total variation norm.
\end{definition}
Definition \ref{definition:stability} is the standard notion of ergodicity of Markov chains, namely convergence to stationarity in total variation \citep{meyn2012markov}. Regardless of the initialization, a stable emulator converges to a unique stationary distribution~$\mu$, which defines its long-run statistics. Long-run fidelity can then be assessed by comparing $\E_\mu[\varphi(X)]$ with the corresponding data statistics $\E_{\mu_\star}[\varphi(X)]$, for suitable observables $\varphi:\R^d\to\R^k$.

\paragraph{Perturbation response.} A useful emulator should also allow us to probe the system's response to an external forcing, or perturbation, as in counterfactual experiments such as the climate's response to a heating anomaly. The forced response is therefore a physically meaningful validation criterion, complementary to stationary and forecasting metrics \citep{majdaSIAM,FabriCoarseGraining}.

\begin{definition}[Perturbations]
    \label{definition:perturbations}
    \begin{sloppypar}
        We define a perturbation kernel parameterized by a vector~$v \in \R^d$ as follows:~$\nu P_v$ is defined as the distribution of $X_{1}$ when~${X_0 = Z + v}$, with $Z \sim \nu$. Equivalently,~${P_v(x, \cdot) := P(x+v, \cdot)}$, and, in particular,~$P_0 = P$.
        Given a perturbation sequence~$u = (u_t)_{t \geq 0}$ of vectors of $\R^d$, the perturbed chain $(X^u_t)$ is the Markov chain defined by~$X^u_{t+1} \sim P_{u_t}(X^u_t, \cdot)$.
    \end{sloppypar}
\end{definition}
Definition \ref{definition:perturbations} incorporates an additive external forcing into an emulator trained on the equilibrium dynamics, through an operator splitting: the unperturbed model is applied to the state shifted by the forcing integrated over one time step (see Appendix \ref{section:forcing}).
This setting covers the classical response experiments of climate science, in which a prescribed tendency is added to the model equations, such as heating and momentum anomalies in atmospheric models \citep{GRITSUN,Pedram1} and surface flux anomalies in ocean models \citep{gregory2016fafmip}.

\paragraph{Problem statement.}
This work aims at developing a generative model that accurately learns the stochastic dynamics from data, while producing stable trajectories in the sense of Definition \ref{definition:stability}. The emulator is evaluated by comparing its stationary distribution $\mu$ to the data distribution $\mu_\star$, and its statistical response to perturbations $\E[\varphi(X^u_t)]$ to the true system's forced response.

\subsection{Stochastic interpolants for probabilistic emulation}

\label{sec:stochastic_interpolants}
Several classes of generative models are available for learning $p(\cdot \mid x)$ from data, such as conditional diffusion models \citep{ho2020denoising, song2020score} and flow matching \citep{lipman2022flow}. Here, we will build on the stochastic interpolant framework~\citep{albergo2025stochastic,pmlr-v235-chen24n}.
Since the kernel is time-homogeneous, we present the mathematical framework for the generation of $X_1$ conditioned on $X_0=x_0 \in \R^d$, to simplify notations. The generative process then applies identically to $X_{t+1}$ conditioned on $X_t = x_t$ for any $t\in \N$.

\paragraph{Generative model.}
\begin{sloppypar}
    Stochastic interpolants model the conditional distribution~$X_1 | X_0=x_0$ through the following stochastic process initialized at~$X_0=x_0$:
    \begin{equation} \label{eq:generative_model_process}
        \begin{aligned}
            \ud X_s & =b(X_s, s; x_0)\, \ud s+\sigma_s\, \ud W_s
            \quad \text{for } s \in [0,1],
        \end{aligned}
    \end{equation}
    where $W_s$ denotes a Wiener process, $\sigma : [0,1] \to \R_+$ is a noise schedule, and the drift $b(x, s; x_0)$ is learned from data, as a function of the triplet~${(x, s ; x_0) \in \R^d \times [0, 1]\times \R^d}$.
    We stress that $s \in [0, 1]$ is a continuous sampling time, different from the physical time index $t \in \N$.
    The output of the generative model is the terminal distribution $p(X_1|X_0=x_0)$ of~\eqref{eq:generative_model_process}, obtained by integrating the stochastic differential equation.
\end{sloppypar}

\paragraph{Training.}
\begin{sloppypar}
    Stochastic interpolants~\eqref{eq:generative_model_process} are motivated by the existence of an optimal drift $b_\star$ that samples the data distribution, and can be learned from data. Define the interpolant process~${{I}_s := {\alpha}_s Y_0 + {\beta}_s Y_1 + {\sigma}_s W_s}$ and~${\dot{I}_s := \dot{\alpha}_s Y_0 + \dot{\beta}_s Y_1 + \dot{\sigma}_s W_s}$, with $(Y_0, Y_1)   \sim \mu_\star(y_0)p_\star(y_1|y_0)$ consecutive states from the data distribution, and coefficients $(\alpha_s)$ and $(\beta_s)$ satisfying the boundary conditions $I_0 = Y_0$ and~$I_1 = Y_1$. Then, \citet[Theorem 3.1]{pmlr-v235-chen24n} show that the conditional expectation~$b_\star(x, s; x_0) := \E[\dot{I}_s | I_0=x_0, I_s=x]$ produces a generative process \eqref{eq:generative_model_process} that samples from the data conditional distribution
    $X_1 | [X_0 =x_0] \sim p_\star(\cdot \mid x_0)$.
    Crucially, the $L^2$-optimality of the mean squared error predictor implies that $b_\star$ is equivalently characterized as the minimizer of the regression loss
    \begin{equation} \label{eq:training_loss}
        {L}(b) = \int_0^1 \mathbb{E} \left[ \big\lVert b(I_s, s; Y_0) - \dot{I}_s \big\rVert^2 \right] \ud s,
    \end{equation}
    where the expectation is taken over consecutive pairs $(Y_0, Y_1)$ of the data distribution,
    and an independent Brownian path $W_s$. Leveraging this regression interpretation, $b_\star$ can be approached by a parametric model trained from data by minimizing \eqref{eq:training_loss} with the stochastic gradient descent algorithm, using batches of consecutive pairs $(y_t, y_{t+1})$ sampled from $\mathcal{D}$.
\end{sloppypar}

Once trained, the obtained drift can directly be used to sample from the stochastic dynamics~\eqref{eq:emulated_trajectory} by integrating~\eqref{eq:generative_model_process}, thus defining an autoregressive stochastic emulator.

\paragraph{Challenges.}
The conditional stochastic interpolant is guaranteed to reproduce the conditional data distribution in the limit of infinite data and expressivity. In practice, data and expressivity are limited, and the learned drift carries approximation errors that accumulate along the rollout, empirically leading to instability. In the next section, we propose an architectural constraint on the drift that provably enforces stability, while preserving learning flexibility.

\section{Forced-dissipative stochastic interpolants} \label{sec:architecture}

In this section, we introduce forced-dissipative stochastic interpolants, a novel stochastic interpolant architecture that enforces stability by design. We draw inspiration from fluid dynamics and derive a constrained drift architecture, provably bounding the energy flow throughout the generative process.

Different techniques can be used to impose physical structure in an emulator, such as a soft penalty in the training loss \citep{raissi2019physics,Nikolaj}, or as a hard constraint in the sampling algorithm \citep{christopher2024projected,blanke2026strictly}. However, these constraints must often be specified in advance, for instance as the energy level of the unforced system, and are thus not directly compatible with an external forcing. We instead impose a hard constraint on the architecture of the drift, that bounds the rate at which energy enters the generative process without fixing the energy level. It requires no knowledge of the equations of the modeled system.


\paragraph{Forced-dissipative drift.}
\begin{sloppypar}
    Inspired by the energy conservation principles of turbulent dynamical systems~\citep{MajdaIntroTurbulence}, we propose a forced-dissipative functional structure of stochastic interpolants.
    Specifically, we parameterize the drift $b$ of the stochastic interpolant process~\eqref{eq:generative_model_process}, function of the triplet~${z:=(x, s; x_0) \in \R^d \times [0,1] \times \R^d}$, as:
    \begin{equation}\label{eq:dissipative_drift}
        b(z) =
        F(z)
        +
        A(z) x
        +
        D(z)x
        +
        S(z)x
    \end{equation}
    where the parameterization enforces, pointwise in $z = (x, s; x_0)$,
    \begin{equation}\label{eq:structural_properties}
        A^\top = -A,
        \qquad
        D \preceq -\gamma I_d,
        \qquad
        \langle y, S\, y \rangle \leq 0 \quad \forall y \in \R^d,
    \end{equation}
    for a fixed dissipation rate $\gamma > 0$. In \eqref{eq:dissipative_drift}, $F$ has the interpretation of an internal forcing term that injects energy, $A$ and $D$ are linear energy-preserving and dissipative terms, and $S$ is a nonlinear energy-preserving term. Such constraints can be enforced by standard linear algebra, \textit{e.g.} through the parameterizations~${A = J - J^\top}$,~$D = -B B^\top - \gamma I_d$ with unconstrained matrices $J, B$, and $S$ can be built as a combination of skew-symmetric matrices with state-dependent weights. The $\gamma$ coefficient is the dissipation rate guaranteed by the architecture. All four components are parameterized by a neural network, which is trained by minimizing the training loss \eqref{eq:training_loss} under this hard architectural constraint. The value of $\gamma$ is chosen empirically, and we found that $\gamma =1$ works well in general, while too large values affect sampling accuracy. More details are given in Appendix \ref{section:experimental_details}.

\end{sloppypar}
\paragraph{Constrained energy flow.}
The purpose of our forced-dissipative stochastic interpolant drift \eqref{eq:dissipative_drift} is to ensure that energy, here defined by the squared norm $\Vert x \Vert^2$, flows in a controlled way. Specifically, the structural properties~\eqref{eq:structural_properties} guarantee that energy can only enter the interpolant process through the forcing term and the noise. The energy production is balanced by the dissipation term, and the net energy variation is controlled by the ratio of their magnitude. This intuition is formalized in the remainder of this section, where we state the central result capturing our architecture's energy structure under regularity assumptions.


\begin{assumption}[Bounded coefficients]\label{assumption:bounded_forcing}
    The forcing term and the noise are uniformly bounded
    \begin{equation}
        \begin{aligned}
            \Vert F \Vert_\infty & := \sup \{ \Vert F(x, s; x_0) \Vert, x, x_0 \in \R^d, \, s \in [0,1]\} < \infty,
            \\
            \sigma_{\max}        & := \sup_{s \in [0,1]} \sigma_s < \infty,
        \end{aligned}
    \end{equation}
    and the three operators are uniformly bounded in operator norm: there exists~$K < \infty$ such that
    \begin{equation}
        \Vert A(z) \Vert + \Vert D(z) \Vert + \Vert S(z) \Vert \leq K
        \qquad \text{for all } z = (x, s; x_0).
    \end{equation}
\end{assumption}

\begin{assumption}[Regular coefficients]\label{assumption:regularity}
    The maps $F, A, D, S$ are continuous on~$\R^d \times [0,1] \times \R^d$ and locally Lipschitz in $(x, x_0)$ uniformly in $s$.
    The noise schedule $\sigma : [0,1] \to \R_+$ is continuous.
\end{assumption}

Under the previous assumptions, our generation process \eqref{eq:generative_model_process} is well posed, and its terminal law~$P(x_0, \cdot)$ defines a Markov kernel (see Appendix \ref{section:proofs}).
By design, our architecture enforces a controlled expected energy rate, as quantified by the following result.
\begin{restatable}[Energy bound]{proposition}{energybound}
    \label{proposition:energy_bound}
    Under the drift structure \eqref{eq:dissipative_drift}--\eqref{eq:structural_properties} and Assumptions \ref{assumption:bounded_forcing} and \ref{assumption:regularity}, for every~$x_0 \in \R^d$, the expected energy $s \mapsto \E[\Vert X_s \Vert^2]$ of the generation process \eqref{eq:generative_model_process} is finite and absolutely continuous on $[0,1]$, and for almost every $s \in [0,1]$,
    \begin{equation}
        \label{eq:expected_energy_bound}
        \frac{1}{2} \frac{\ud }{\ud s} \E[\Vert X_s \Vert^2]
        \leq
        - \frac{\gamma}{2} \E[\Vert X_s \Vert^2] + \frac{1}{2\gamma} \Vert F \Vert_\infty^2 + \frac{1}{2}d \sigma_s^2.
    \end{equation}
\end{restatable}
\paragraph{Applicability.}
Note that, although inspired by fluid dynamics, our generative model applies to dynamical systems and data of arbitrary nature: the energy-dissipative structure and the forcing component are internal to the neural network's architecture, and do not require any assumption on the modeled system.

In the next section, we show how Proposition \ref{proposition:energy_bound} results in a practical stability guarantee for the emulator, relying on the Foster--Lyapunov formalism for Markov chains.

\section{Stability guarantees} \label{sec:analysis}

In this section, we derive stability guarantees for our forced-dissipative stochastic interpolant architecture, for free rollouts and perturbed regimes. These guarantees are established within the classical Markov chain stability framework. Specifically, we derive a Foster--Lyapunov condition, which allows us to prove geometric ergodicity. Our analysis is built on the framework of \citet{meyn2012markov}. In the remainder of this section, we assume the drift structure~\eqref{eq:dissipative_drift}--\eqref{eq:structural_properties}.

We first require a nondegeneracy condition on the noise schedule, as a technical detail allowing us to prove that the Markov chain is irreducible.
\begin{assumption}[Nonvanishing noise]\label{assumption:nonvanishing_noise}
    There exists $\sigma_{\min} > 0$ such that $\sigma_s \geq \sigma_{\min}$ for all $s \in [0, 1]$.
\end{assumption}
Assumption \ref{assumption:nonvanishing_noise} is a technical requirement for the following results. Note that it deviates from the ideal stochastic interpolant scheme, which requires $\sigma_1 = 0$. However, this floor can be taken as small as desired, in particular smaller than the discretization limit of \eqref{eq:generative_model_process}, so that it plays no role in practice.

Our analysis builds on the following Lyapunov function measuring the system's energy:
\begin{equation}
    V(x) := 1 + \Vert x \Vert^2.
\end{equation}
The energy bound \eqref{eq:expected_energy_bound}, integrated over one generation step, yields the following Foster--Lyapunov condition for $(X_t)$.
\begin{restatable}[Foster--Lyapunov drift condition]{proposition}{fosterlyapunov}
    \label{proposition:foster}
    Under Assumptions \ref{assumption:bounded_forcing} and~\ref{assumption:regularity}, the emulated chain \eqref{eq:emulated_trajectory} satisfies the Foster--Lyapunov condition
    \begin{equation}
        \label{eq:Foster_Lyapunov}
        \E
        \left[
            V(X_{t+1})
            \mid X_t=x
            \right]
        \leq
        a V(x)+c\quad \text{for all} \; x \in \R^d,
    \end{equation}
    with  $a := e^{-\gamma} \in (0,1)$ and $c := (1 - e^{-\gamma}) \big( 1 + \Vert F \Vert_\infty^2/\gamma^2 + d \, \sigma_{\max}^2/\gamma \big) \in (0, \infty)$.
\end{restatable}

Proposition \ref{proposition:foster} shows that the emulator's energy cannot grow too far from an equilibrium, as it is constrained by a downward drift in expectation.
The next result formalizes how this Foster--Lyapunov inequality leads to a formal stability statement for our forced-dissipative emulator.
\begin{restatable}[Geometric ergodicity]{proposition}{ergodicity}
    \label{proposition:ergodicity}
    Under the drift structure \eqref{eq:dissipative_drift}--\eqref{eq:structural_properties} and Assumptions \ref{assumption:bounded_forcing}--\ref{assumption:nonvanishing_noise}, the emulated chain \eqref{eq:emulated_trajectory} admits a unique invariant probability measure $\mu$, which has finite second moments.
    Moreover, there exist $M < \infty$ and $\rho \in (0, 1)$ such that the chain converges to $\mu$ at the geometric rate $\rho$:
    \begin{equation}
        \label{eq:geometric_ergodicity}
        \big\Vert P^t(x, \cdot) - \mu \big\Vert_{\mathrm{TV}} \leq M \rho^t \, V(x)
        \qquad \text{for all } x \in \R^d \text{ and } t \in \N .
    \end{equation}
    In particular, the emulated chain is stable in the sense of Definition \ref{definition:stability}.
\end{restatable}
The proof in Appendix \ref{section:proofs} establishes the same bound for the stronger $V$-norm, which also controls the unbounded observables of at most quadratic growth, such as the mean and the energy.

The results above concern the unperturbed chain. Since the purpose of an emulator is to study responses to external forcings, it is crucial that stability guarantees extend to the perturbed dynamics. For a general, non-stationary bounded sequence of perturbations, we first show that our forced-dissipative structure constrains the chain to stay bounded in energy.
\begin{sloppypar}
    \begin{restatable}[Bounded energy under time-varying perturbations]{proposition}{forcedstability}
        \label{proposition:forced_bounded}
        Assume Assumptions \ref{assumption:bounded_forcing}--\ref
        {assumption:regularity}. Let~${u = (u_t)_{t \geq 0}}$ be a perturbation sequence of bounded amplitude, $\sup_{t} \Vert u_t \Vert < +\infty$, and let $(X_t^u)$ be the perturbed chain, started from an initial condition $X^u_0$ of finite second moment. Then the second moment~$\E \big[ \Vert X_t^u \Vert^2 \big]$ is finite for every $t \in \N$ and bounded uniformly in time.
    \end{restatable}
\end{sloppypar}

Finally, we state an ergodicity result in the case of steady perturbations.

\begin{restatable}[Perturbed geometric ergodicity]{proposition}{forcedergodicity}
    \label{proposition:forced_ergodicity}
    Under the assumptions of Proposition \ref{proposition:ergodicity}, let the perturbation be constant in time: $u_t = v$ for all $t \in \N$, for some $v \in \R^d$. Then the time-homogeneous perturbed chain $(X^v_t)$ is geometrically ergodic, in the sense of Proposition \ref{proposition:ergodicity}.
\end{restatable}
Proposition \ref{proposition:forced_ergodicity} guarantees that, under arbitrarily large constant perturbations, the emulator remains stable and converges to a perturbed stationary distribution. In particular, the emulator does not blow up under large steady forcings, which are common in climate experiments~\citep{falasca2026physics}.

The results above guarantee the theoretical stability of our architecture only in the continuous limit of~\eqref{eq:generative_model_process}, and they do not prove anything about its learning accuracy. In the next section, we validate our forced-dissipative model empirically.

\section{Application to the emulation of two-scale dynamics} \label{sec:experiments}

In this section, we evaluate the emulation ability of our forced-dissipative architecture for a turbulent dynamical system, in various settings. We compare it to other generative models, including simpler baselines that are stable by design. More details about the physical systems, the models and the evaluation protocols in the experiments are provided in Appendix \ref{section:experimental_details}.

\paragraph{Physical systems.}
We evaluate the emulators on two modified versions of the Lorenz-96 system~\citep{Lorenz96}, a standard benchmark for multiscale turbulent dynamical systems in atmospheric sciences, with $d$ slow variables and $m$ fast variables. We first consider the simpler single-layer system, where $m=0$, in which the emulator observes the full state. We then consider an anisotropic two-layer system, with $m=10$, in which the emulator observes only the slow variables, while the unobserved fast variables act on them as an effective noise.
We generate a dataset of $10^5$ consecutive states in $\R^{d \times (m+1)}$, that we use as a training set for the compared data-driven emulators.


\paragraph{Evaluation.}
For both systems, we evaluate the long-run behavior of the emulators with two criteria, compared against the truth. We first compute the stationary distribution obtained by integrating a long rollout. We then compute the response of the emulator to large external perturbations, measured as the statistical response operator
\begin{equation}
    R_t(u, \varphi) :=
    \E \left[
        \varphi(X^u_t)
        \right]
    -
    \E_\mu \left[
        \varphi(X)
        \right],
    \qquad X^u_0 \sim \mu,
\end{equation}
where $(X^u_t)$ is the perturbed chain of Definition \ref{definition:perturbations} and $\mu$ is the stationary distribution. In this work, we focus on the responses of the mean and of the variance, that is $\varphi(x) = x$ and $\varphi(x) = \Vert x - \E_\mu[x]\Vert^2$. The statistical response is computed by a Monte Carlo estimation over a statistical ensemble, and compared to the truth. The external forcing is added to the right-hand side of the true dynamics, and it enters the emulators as the perturbation sequence of Definition~\ref{definition:perturbations} given by the forcing integrated over one observation interval.

\subsection{Fully-observed dynamics}
\label{section:single_layer}

\begin{figure}[t]
    \begin{subfigure}[b]{.15\linewidth}
        \centering
        \includegraphics[height=3cm, trim={.2cm .2cm 0 .2cm}, clip]{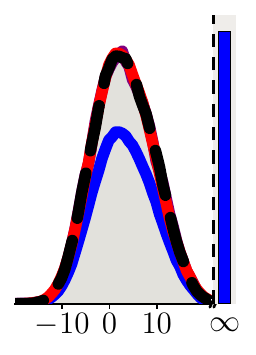}
    \end{subfigure}
    \begin{subfigure}[b]{.1\linewidth}
        \centering
        \includegraphics[height=1.2cm, trim={.2cm .0cm 0 0cm}, clip]{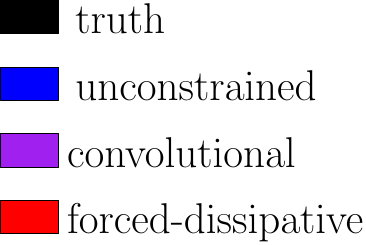}
        \par \bigskip
        \par \bigskip
    \end{subfigure}
    \quad
    \begin{subfigure}[b]{.6\linewidth}
        \centering
        \includegraphics[width=\linewidth, trim={0 .2cm 0 .2cm}, clip]{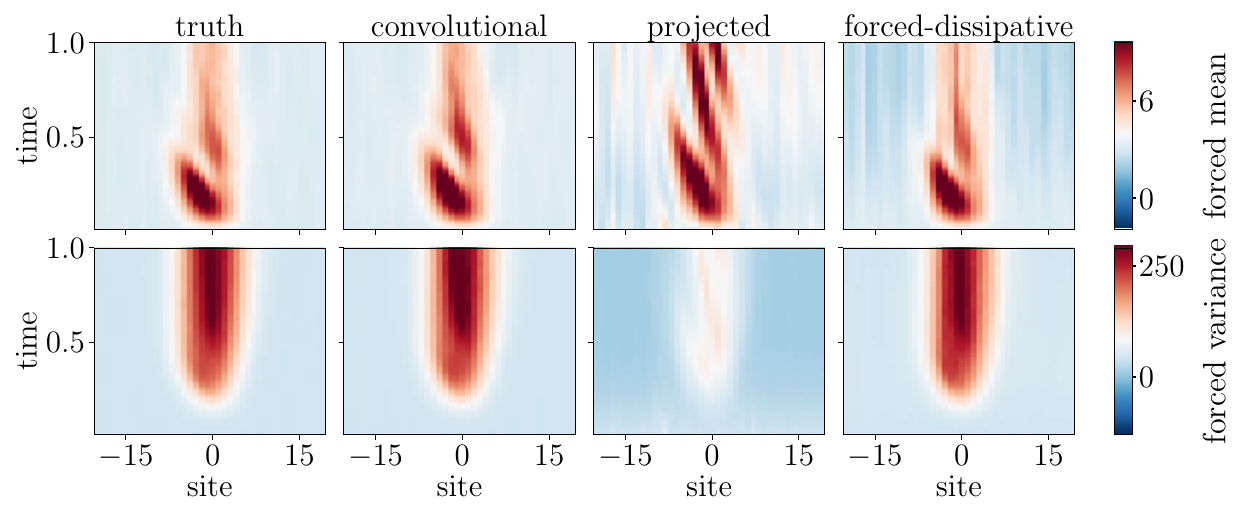}
    \end{subfigure}
    \caption{Stationary distributios and response to large perturbations for the single-layer system.}
    \label{figure:single_layer}
\end{figure}

We first consider the single-layer Lorenz-96 system, with $d=40$ variables on a periodic ring and a uniform forcing in the chaotic regime. The generative models are trained on the full, single-layer state. Since the forcing is uniform, the dynamics are invariant under translations along the ring.

\paragraph{Compared methods.}
Our forced-dissipative stochastic interpolant is compared with stochastic interpolant baselines, which all share the same interpolant, the same training loss \eqref{eq:training_loss} and the same sampler, and differ only in the parameterization of the drift.
An {unconstrained interpolant} parameterizes the drift with a fully-connected neural network, without any structural constraint.
A {projected interpolant} uses the same unconstrained neural network, but rescales the state to the mean energy of the training data at each sampling step, which is a straightforward strategy for imposing a bounded energy.
A {convolutional interpolant} parameterizes the drift with a convolutional neural network (CNN). Its locality and its translation equivariance are symmetries of the Lorenz-96 dynamics, so that it serves as an informed oracle baseline that is given the structure of the system.
The {forced-dissipative interpolant} is our architecture~\eqref{eq:dissipative_drift} of Section~\ref{sec:architecture}.

\paragraph{Experimental setup.}
We first evaluate the emulator's stationary distribution. Each emulator is run for $10^6$ steps to compute histograms of its marginals, compared with that of a true trajectory.
We then evaluate the response to a large steady forcing, localized on a few sites of the ring. Its amplitude is 10 times the standard deviation of the stationary distribution, far beyond the linear response regime.

\paragraph{Results.}
The results are shown in Figure \ref{figure:single_layer}. While the unconstrained stochastic interpolant approaches the correct stationary distribution, it ultimately diverges, and thus cannot reliably emulate the system. The~CNN, the projected interpolant, and the forced-dissipative interpolant all learn the distribution accurately.
Regarding the statistical responses, the forced-dissipative interpolant responds as well as the~CNN interpolant, which has a strong physical prior.
The unconstrained stochastic interpolant systematically goes unstable. While the projected interpolant remains stable by design, its response is inaccurate, because imposing a fixed energy in the generative model is incompatible with perturbations, which change energy by definition.

\subsection{Partially observed two-scale dynamics}
\label{section:two_layer}

\begin{figure}[t]
    \begin{subfigure}[b]{.37\linewidth}
        \centering
        \includegraphics[height=.7cm]{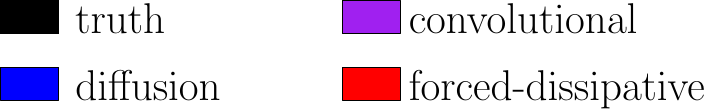}
        \par
        \includegraphics[width=.7\linewidth, trim={0 .2cm 0 .2cm}, clip]{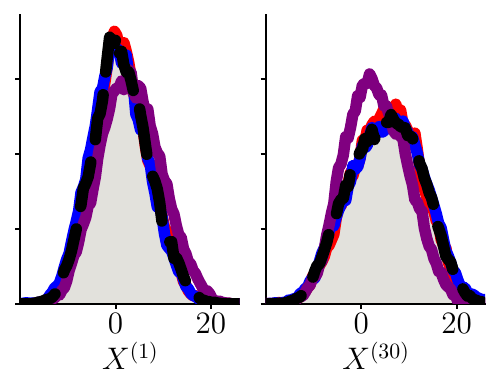}
    \end{subfigure}
    ~
    \begin{subfigure}[b]{.6\linewidth}
        \includegraphics[width=\linewidth, trim={0 .2cm 0 .2cm}, clip]{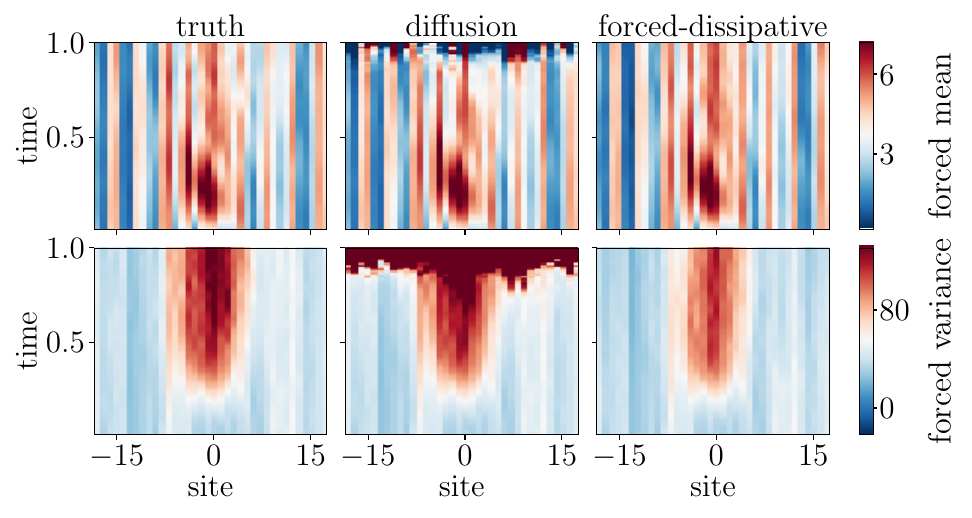}
    \end{subfigure}
    \caption{Stationary distributions and response to large perturbations for the two-layer system.}
    \label{figure:two_layer}
\end{figure}
We now consider the more realistic, partially observed setting
in which each of the $36$ slow variables on a ring is coupled to $m=10$ fast variables that evolve ten times faster. The truth system thus evolves in dimension~$396$. The emulators observe only the $36$ slow variables, so that the unobserved fast variables act on them as an effective noise. The forcing of the slow variables differs from site to site. Each site therefore has its own stationary distribution, and the system is no longer invariant under translations along the ring. This modified Lorenz-96 dynamics was introduced by \citet{falasca2026physics} and exhibits a more complex, spatially heterogeneous turbulent regime, in contrast with the spatially homogeneous turbulence of the original system (see Appendix \ref{section:experimental_details}).

\begin{wrapfigure}[22]{right}{4.2cm}
    \hfill\includegraphics[height=.25cm]{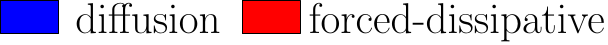}
    \par
    \includegraphics[width=\linewidth, trim={0 .2cm, 0 , .5cm}, clip]{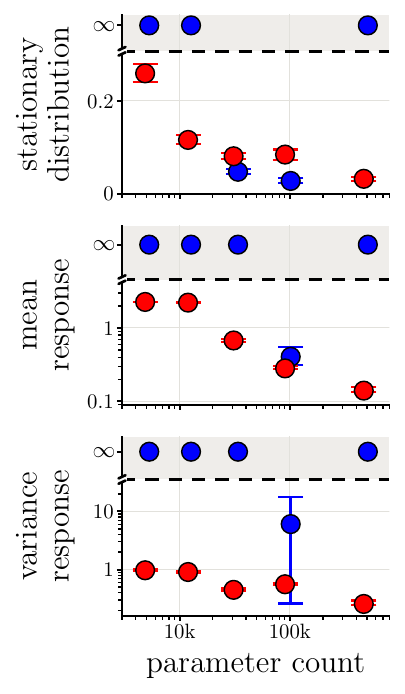}
    \caption{Emulation error as a function of the generative model's capacity. The~$\infty$ symbol denotes rollout instability.}
    \label{figure:capacity}
\end{wrapfigure}

\paragraph{Compared methods.}
The unconstrained emulators of Section \ref{section:single_layer} were systematically unstable or inaccurate in this experiment, so they are not included. Instead, a {conditional diffusion model} \citep{ho2020denoising} is implemented, with a fully-connected neural network for the denoiser. It represents a standard generative emulator of a different type, often used for fluid emulation \citep{rozet2026lost}.

\paragraph{Experimental setup.}
The same experimental setup of the previous section is adopted, with a smaller $5 \sigma$ forcing to ensure the diffusion model stays stable.
In order to study the methods' stability, we study how the stationary and response errors depend on the capacity of the neural network. The diffusion model and the forced-dissipative interpolant are trained until convergence on the same dataset at parameter counts ranging over two orders of magnitude. We report the Wasserstein-1 distance between the emulated and true marginal distributions, averaged over the slow variables and divided by the standard deviation of the stationary distribution. The response errors are the relative quadratic errors of the response in mean and variance of the slow variables over the first $50$ steps, averaged over 5 ensembles. Both errors are set to infinity when a single rollout of the emulator diverges.
Finally, we apply a time-dependent forcing, with the same spatial profile modulated in time by a sinusoid and an amplitude rising linearly, as a proxy for the seasonal cycle of climate forcings. We compute the emulators' response to this forcing for the best-performing models of each category.

\begin{wrapfigure}[18]{right}{4.cm}
    \hfill\includegraphics[height=.6cm]{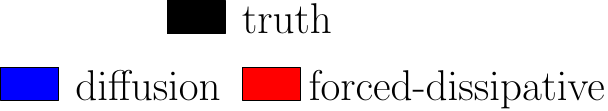}
    \includegraphics[width=\linewidth, trim={0 .2cm, 0 , .5cm}, clip]{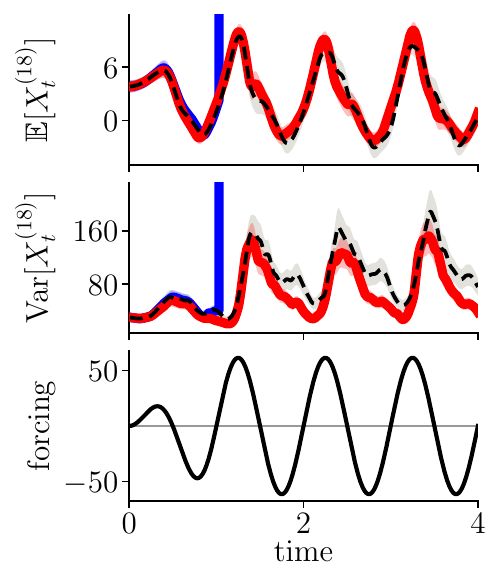}
    \caption{Mean and variance at the forcing site under an oscillating forcing.}
    \label{figure:oscillatory}
\end{wrapfigure}
\paragraph{Results.}
The stationary distribution and responses of the best-performing emulators are shown in Figure~\ref{figure:two_layer}.
The convolutional stochastic interpolant is not suited for this experiment, since it cannot reproduce this system's anisotropy. Both the diffusion model and our emulator reproduce the stationary statistics faithfully. Our forced-dissipative interpolant responds accurately, and remains stable under the steady forcing, as stated in Proposition \ref{proposition:forced_ergodicity}. The diffusion model does not diverge but produces unphysically large responses.
The stability study, in Figure \ref{figure:capacity}, reveals that the diffusion model is unstable in the forced regime below 32k parameters and unstable in free rollouts below 10k parameters. Strikingly, its stability is not guaranteed by a more capable model, as the diffusion model of 500k parameters is unstable, although it has the smallest validation loss among all diffusion models. The diffusion model with 100k parameters is the best one, with the lowest distribution error but an unreliable response error. In contrast, our architecture remains stable regardless of the neural network's capacity, as opposed to the conditional diffusion model. This is crucial as emulators are typically trained in a low data regime for climate applications, thus severely limiting the model's capacity.
The difference is even stronger under the oscillating forcing (Figure \ref{figure:oscillatory}), where the diffusion model starts to go unstable as the forcing first reaches its full amplitude. The forced-dissipative interpolant tracks the true mean response and stays bounded, in agreement with Proposition \ref{proposition:forced_bounded}. The accuracy in variance is initially good then decreases (see Appendix \ref{section:additional-experimental-results} for more details).

\section{Related work} \label{sec:related_work}

\paragraph{Probabilistic forecasting.}
Conditional generative models are now standard for probabilistic emulation of partially observed dynamics, through stochastic interpolants, flow matching and diffusion models \citep{pmlr-v235-chen24n, NEURIPS2025_607b356c, rozet2026lost, kohl2023acdm, ruhlingcachay2024dyffusion, ProbabilisticSIPedram, accarino2026wavelet}.
While the stability of a single generation pass has been analyzed \citep{strasman2026forgetting}, the long-run behavior of the resulting autoregressive Markov chain remains largely unexplored.
\citet{pathak2026engression} prove geometric ergodicity of a closed-loop generative forecaster where the stability condition is assumed rather than imposed.


\paragraph{Constrained neural emulators.}
Stability by construction has been studied for deterministic neural models, by learning a Lyapunov function jointly with the dynamics \citep{manek2019learning, perez2026let}, by encouraging or projecting the dynamics onto a dissipative set \citep{li2021learning, tang2024dissipative, goertzen2025eco}, and by writing energy structure into the architecture~\citep{desai2021port, roth2026stable, gruber2025metriplectic, VANGASTELEN2024113003,PierreConstraints}. These guarantees do not extend to probabilistic models, which call for the Markov chain framework instead. For generative emulators, structure has been imposed through energy-consistent interpolant coefficients \citep{Nikolaj}, conservative-dissipative flow matching \citep{baheri2025metriplectic}, or projection in the sampling loop \citep{christopher2024projected, blanke2026strictly}. These approaches constrain the system to a predetermined energy. They are thus not directly suitable for perturbed dynamics, where the system is forced to go out of equilibrium.

\section{Discussion}

We introduced forced-dissipative stochastic interpolants, a generative architecture whose drift has controlled energy flow. This constraint is shown to imply geometric ergodicity for our emulator. We demonstrate in experiments that these results apply in practice, where unconstrained generative models diverge, including under large forcings, and that our model accurately predicts the perturbation responses of external forcings.

\paragraph{Limitations and future work.}
Our guarantees hold for the continuous-time generative process, and the Foster--Lyapunov criterion could be extended to the discrete-time sampler. Furthermore, the matrix products of our parameterization become costly in high dimension, so that the method is best suited to reduced-order modeling in a low-dimensional latent space \citep{rozet2026lost}.
The Foster--Lyapunov condition is a design principle that is not specific to stochastic interpolants. It could be imposed on cheaper generative models, such as discrete-time models \citep{falasca2026physics}.
In future work, we will apply the method to more complex systems, such as quasi-geostrophic turbulence \citep{salmon1978two}.

\clearpage
\paragraph{Reproducibility statement.}
Experimental details are provided in Appendix \ref{section:experimental_details}.
The proofs of the theoretical results included in this paper are available in Appendix \ref{section:proofs}.
\clearpage

\bibliographystyle{unsrtnat}
\bibliography{references}

\clearpage
\newpage
\appendix

\input{appendix.tex}

\end{document}

%% file: appendix.tex
\section{Extended related work} \label{section:extended_related_work}

This appendix expands the discussion of Section \ref{sec:related_work}.

\paragraph{Forced-dissipative dynamics.}
Forced-dissipative structure is the classical setting of turbulent dynamics \citep{MajdaBook, majda2016introduction}, and the balance between forcing and dissipation is what is used to prove ergodicity for stochastic differential equations and for the stochastic Navier--Stokes equations \citep{mattingly2002ergodicity, goldys2005exponential}.

\section{External forcing in the emulator} \label{section:forcing}

The emulators are trained on unforced trajectories only, and the forcing is not an input of the network. It enters at rollout time through an operator splitting. Write the forced dynamics of the observed variables as $\dot x = f(x) + G(t)$, where $f$ is the unforced dynamics. Over one observation interval $[t\tau, (t+1)\tau]$, we split it into the forcing alone, $\dot x = G(t)$, followed by the unforced dynamics $\dot x = f(x)$. This is the Lie--Trotter splitting \citep{mclachlan2002splitting}. It approximates the forced dynamics, and the approximation improves as the observation interval $\tau$ decreases. The flow of the forcing alone is the shift $x \mapsto x + u_t$ by the integrated forcing
\begin{equation}
    u_t = \int_{t \tau}^{(t+1) \tau} G(s) \ud s ,
\end{equation}
and the law of the unforced dynamics over one interval is what the kernel $P$ learns. One step of the forced emulator is therefore
\begin{equation}
    X^u_{t+1} \sim P(X^u_t + u_t, \cdot) = P_{u_t}(X^u_t, \cdot),
\end{equation}
which is the perturbed chain of Definition~\ref{definition:perturbations}. For a steady forcing, $u_t = \tau G$.

\section{Experimental details} \label{section:experimental_details}

\subsection{Dynamical systems} \label{app:systems}

\paragraph{Single-layer Lorenz-96.}
The single-layer Lorenz-96 system \citep{Lorenz96} evolves $d = 40$ variables $x_0, \dots, x_{d-1}$ on a periodic ring, with indices taken modulo $d$, as
\begin{equation}
    \frac{\ud x_k}{\ud t} = x_{k-1} \big( x_{k+1} - x_{k-2} \big) - x_k + F,
    \qquad k = 0, \dots, d-1,
\end{equation}
with the uniform forcing $F = 16$. The quadratic term conserves the energy $\frac12 \Vert x \Vert^2$, the linear term dissipates it and $F$ injects it, which is the forced-dissipative structure that our architecture reproduces. The system is integrated with the fourth-order Runge--Kutta scheme with time step $0.01$, from a state near the fixed point $x_k = F$. After a burn-in of $10^5$ steps, states are recorded every $\tau = 0.01$ time units, and consecutive recorded states form the transition pairs. The standard deviation of the stationary distribution, $\sigma = \big( \frac1d \sum_k \mathrm{Var}(x_k) \big)^{1/2}$, is $\sigma = 6.32$.

\paragraph{Two-layer Lorenz-96.}
The two-layer system \citep{Lorenz96} 
couples $d = 36$ slow variables $X_k$ to $m = 10$ fast variables $Y_{j,k}$ per slow variable,
\begin{align}
    \frac{\ud X_k}{\ud t}     & = X_{k-1} \big( X_{k+1} - X_{k-2} \big) - X_k + F_k - \frac{h_x}{m} \sum_{j=1}^{m} Y_{j,k},          \\
    \frac{\ud Y_{j,k}}{\ud t} & = \frac{1}{\varepsilon} \Big( Y_{j+1,k} \big( Y_{j-1,k} - Y_{j+2,k} \big) - Y_{j,k} + h_y X_k \Big),
\end{align}
with $h_x = h_y = 1$ and the time-scale separation $\varepsilon = 0.1$. The slow variables are periodic in $k = 0, \dots, d-1$. The fast variables form a single ring of $m d$ variables, so that $Y_{m+1,k} = Y_{1,k+1}$. The forcing is drawn once from $F_k \sim \mathcal{N}(16, 10^2)$, redrawing any value below $4$, which gives $F_k \in [5.6, 37.4]$. It is then kept fixed. The system is integrated with the fourth-order Runge--Kutta scheme with time step $0.005$. After a burn-in of $10^4$ recorded states, states are recorded every $\tau = 0.01$ time units. The emulators observe only the slow variables. The standard deviation of the stationary distribution of the slow variables, defined as for the single-layer system, is $\sigma = 6.15$.

\paragraph{External forcings.}
The response experiments add an external forcing $G \in \R^d$ to the right-hand side of the equations of the true system, which are the slow equations in the two-layer system. Its spatial profile is the Gaussian bump
\begin{equation} \label{eq:forcing_profile}
    G_k = w \exp\!\Big( - \frac{\delta(k, k_0)^2}{2 \ell^2} \Big),
\end{equation}
where $\delta(k, k_0)$ is the distance between sites $k$ and $k_0$ along the ring, and $\ell = 3$. The center $k_0$ is $20$ in the single-layer system and $18$ in the two-layer system. The amplitude $w$ is a multiple of $\sigma$ given in Appendix~\ref{app:evaluation}. In the true system, the forcing is evaluated at every stage of the Runge--Kutta scheme.

\subsection{Generative models} \label{app:methods}

\paragraph{Interpolant and sampler.}
All emulators except the diffusion model use the interpolant

\begin{equation}
    I_s = (1-s) X_0 + s X_1 + (1-s) W_s,
\end{equation}
so that $\sigma_s = 1 - s$. The regression target of \eqref{eq:training_loss} is then $\dot I_s = X_1 - X_0 - W_s$. During training, $s$ is drawn uniformly in $[0, 1]$ and $W_s = \sqrt{s}\, z$ with $z \sim \mathcal{N}(0, I_d)$. One step of the emulator integrates \eqref{eq:generative_model_process} from $s = 0$ to $s = 1$ with the Euler--Maruyama scheme on $N_s$ uniform steps of size $\Delta s = 1/N_s$,
\begin{equation}
    X_{s + \Delta s} = X_s + b(X_s, s; x_0)\, \Delta s + (1 - s) \sqrt{\Delta s}\, \xi_s,
    \qquad \xi_s \sim \mathcal{N}(0, I_d).
\end{equation}
The last step is taken at $s = 1 - \Delta s$, so that every step of the discrete sampler injects noise of variance at least $\Delta s^3 > 0$. This positive noise floor plays the role of $\sigma_{\min}$ in Assumption~\ref{assumption:nonvanishing_noise}. The number of sampler steps $N_s$ of each experiment is given in Table~\ref{tab:models}. The time $s$ enters the forced-dissipative interpolant and the diffusion model through the sinusoidal features $\sin(k \pi s)$ and $\cos(k \pi s)$ for $k = 1, \dots, 8$. The unconstrained and convolutional interpolants take $s$ directly as an input.

\paragraph{Unconstrained interpolant.}
The drift is an MLP with SiLU activations and $L$ linear layers of width $w$, applied to the concatenation of $x$, $x_0$ and $s$, where $s$ is repeated $d$ times.

\paragraph{Projected interpolant.}
The projected interpolant uses the weights of the unconstrained interpolant. After every Euler--Maruyama step, the state is rescaled as $X_s \leftarrow X_s \sqrt{\bar E / E(X_s)}$, where $E(x) = \frac12 \Vert x \Vert^2$ and $\bar E$ is the mean of $E$ over the training data. The energy of every intermediate state, and therefore of every emulated state, is then equal to $\bar E$.

\paragraph{Convolutional interpolant.}
The drift is a CNN with SiLU activations and $2$ one-dimensional convolutional layers of kernel size $3$ with circular padding. Its input has three channels, holding $x$, $x_0$ and $s$ at every site, and its hidden layer has $c$ channels. The network is local and equivariant under translations along the ring.

\paragraph{Forced-dissipative interpolant.}
The drift has the structure \eqref{eq:dissipative_drift},
\begin{equation}
    b(x, s; x_0) = F(x_0, s) + \big( A(x_0, s) + D(x_0, s) \big) x + \sum_{i=1}^{M} \phi_i(x - x_0, s)\, A_i(x_0, s)\, x .
\end{equation}
A trunk MLP with $L$ layers of width $w$, RMS normalization and SiLU activations maps $(x_0, s)$ to a hidden representation $h$, and linear heads map $h$ to the operators. The conservative term is $A = U V^\top - V U^\top$ with $U, V \in \R^{d \times r}$. The dissipation is
\begin{equation}
    D = -B B^\top - \gamma I_d ,
\end{equation}
where $B$ is lower triangular with a diagonal passed through a softplus, so that $D \preceq -\gamma I_d$. All forced-dissipative models use $\gamma = 1$. The nonlinear term uses $M$ skew-symmetric matrices $A_i = U_i V_i^\top - V_i U_i^\top$ of rank at most $2r$, weighted by the outputs $\phi_i$ of a second MLP of $(x - x_0, s)$. The MLPs have bounded outputs, using $\mathrm{tanh}$ layers. Since every $A_i$ is skew-symmetric, $\langle x, S x \rangle = 0$, and the constraints \eqref{eq:structural_properties} hold with the dissipation carried by $D$ alone.
The forcing is a bounded neural network, with amplitude bound $10^3$ per coordinate, chosen well above the typical values obtained from the unconstrained network.

\paragraph{Conditional diffusion model.}
The diffusion model uses the variance-preserving cosine schedule $\bar a(s) = \sin^2(\pi s / 2)$, where $s = 0$ is pure noise and $s = 1$ is the data.
It is trained to predict the noise $z$ from $\bar a(s)^{1/2} X_1 + (1 - \bar a(s))^{1/2} z$, $X_0$ and $s$. The prediction is $(1 - \bar a(s))^{1/2} x$ plus the output of an MLP with SiLU activations and $L$ linear layers of width $w$, whose input is the concatenation of $x$, $x_0$ and the features of $s$. The first term is the exact prediction at $s = 0$ and vanishes at $s = 1$. Samples are drawn with the ancestral sampler of \citet{ho2020denoising} on $N_s$ uniform steps, starting from $\mathcal{N}(0, I_d)$. The per-step retention factor $\bar a(s_k) / \bar a(s_{k+1})$ is clipped below at $10^{-3}$.

\paragraph{Training.}
Every network is trained with the Adam optimizer on consecutive pairs of a true trajectory, and the last $10\%$ of the pairs are held out for validation. The models of the single-layer response experiment are trained for $1.5 \times 10^5$ steps with batches of $1{,}024$ pairs, at a constant learning rate of $5 \times 10^{-4}$ for the unconstrained and convolutional interpolants and $5 \times 10^{-5}$ for the forced-dissipative interpolant. The two-layer models of $5.7 \times 10^6$ parameters are also trained at a constant learning rate. 
All other models follow a convergence rule. The learning rate starts at $10^{-3}$ and is divided by $10$ when the mean validation loss over a block of steps improves by less than $0.2\%$, down to $10^{-5}$, where training stops under the same rule with a threshold of $0.1\%$. We keep the weights with the lowest validation loss. The blocks have $2 \times 10^4$ steps with batches of $1{,}024$ pairs for the single-layer system, and $5 \times 10^4$ steps with batches of $128$ pairs for the two-layer capacity study. In each experiment, the unconstrained and forced-dissipative interpolants are matched in parameter count. Table~\ref{tab:models} lists the models of every experiment, and Table~\ref{tab:capacity} the sizes of the capacity study.

The number of sampling steps is chosen large enough to ensure no accuracy is lost.

\begin{table}[h]
    \vspace{1em}
    \centering
    \caption{Emulators of each experiment. $L$ is the number of layers, $w$ the width, $c$ the number of channels, $r$ the rank and $M$ the number of nonlinear modes. $N_s$ is the number of sampler steps per emulator step. All forced-dissipative models use $\gamma = 1$.}
    \label{tab:models}
    \resizebox{\linewidth}{!}{%
        \begin{tabular}{lllrr}
            \toprule
            Experiment                 & Emulator                  & Size                                  & Parameters                             & $N_s$  \\
            \midrule
            Single-layer, stationary   & Unconstrained / projected & $L = 4$, $w = 120$                    & $48{,}400$                             & $100$  \\
                                       & Convolutional             & $c = 128$                             & $1{,}665$                              & $100$  \\
                                       & Forced-dissipative        & $L = 3$, $w = 32$, $r = 2$, $M = 2$   & $48{,}630$                             & $100$  \\
            \midrule
            Single-layer, response     & Unconstrained / projected & $L = 4$, $w = 1254$                   & $3{,}349{,}474$                        & $50$   \\
                                       & Convolutional             & $c = 128$                             & $1{,}665$                              & $50$   \\
                                       & Forced-dissipative        & $L = 4$, $w = 256$, $r = 16$, $M = 8$ & $3{,}347{,}688$                        & $50$   \\
            \midrule
            Two-layer, stationary      & Convolutional             & $c = 16$                              & $209$                                  & $100$  \\
                                       & Diffusion                 & $L = 4$, $w = 1660$                   & $5{,}722{,}056$                        & $100$  \\
                                       & Forced-dissipative        & $L = 4$, $w = 256$, $r = 32$, $M = 8$ & $5{,}724{,}422$                        & $100$  \\
            \midrule
            Two-layer, steady response & Diffusion                 & $L = 3$, $w = 256$                    & $97{,}828$                             & $1600$ \\
            and oscillating forcing    & Forced-dissipative        & $L = 3$, $w = 88$, $r = 7$, $M = 8$   & $487{,}860$                            & $1600$ \\
            \midrule
            Two-layer, capacity        & Diffusion                 & see Table~\ref{tab:capacity}          & $5.1 \times 10^3$ to $4.9 \times 10^5$ & $1600$ \\
                                       & Forced-dissipative        & see Table~\ref{tab:capacity}          & $5.1 \times 10^3$ to $4.9 \times 10^5$ & $1600$ \\
            \bottomrule
        \end{tabular}}
    \vspace{1em}
\end{table}

\begin{table}[h]
    \vspace{1em}
    \centering
    \caption{Sizes of the capacity study. Both architectures have $L = 3$ layers of width $w$.}
    \label{tab:capacity}
    \small
    \begin{tabular}{rrr@{\qquad}rr}
        \toprule
        \multicolumn{3}{c}{Forced-dissipative} & \multicolumn{2}{c}{Diffusion}                                     \\
        $w$                                    & $(r, M)$                      & Parameters  & $w$   & Parameters  \\
        \midrule
        $4$                                    & $(1, 2)$                      & $5{,}064$   & $32$  & $5{,}092$   \\
        $8$                                    & $(2, 3)$                      & $12{,}409$  & $64$  & $12{,}196$  \\
        $16$                                   & $(3, 4)$                      & $32{,}146$  & $128$ & $32{,}548$  \\
        $32$                                   & $(4, 6)$                      & $94{,}216$  & $256$ & $97{,}828$  \\
        $88$                                   & $(7, 8)$                      & $487{,}860$ & $640$ & $490{,}276$ \\
        \bottomrule
    \end{tabular}
    \vspace{1em}
\end{table}

\subsection{Evaluation} \label{app:evaluation}

\paragraph{Stationary distribution.}
Every chain starts from the same true state and runs for $10^6$ steps after a spin-up of $1000$ steps. The true reference is a trajectory of $10^6$ states. The time that diverged chains would have spent after their divergence is counted as probability mass at infinity.

\paragraph{Steady response, single-layer system.}
The forcing \eqref{eq:forcing_profile} has amplitude $w = 10\,\sigma = 63.2$. The ensemble has $1{,}024$ independent members. The forcing is switched on at $t = 0$ and held for $100$ steps. The truth is integrated with Runge--Kutta steps of size $0.01$. At every step and every site, we report the ensemble mean and the ensemble variance over the members that have not diverged, for the truth and for each emulator.

\paragraph{Steady response, two-layer system.}
The two emulators are the best model of each architecture in the capacity study.
The forcing \eqref{eq:forcing_profile} has amplitude $w = 5\,\sigma$, because the diffusion model diverged for larger forcings.

\paragraph{Oscillating response, two-layer system.}
The forcing has the spatial profile \eqref{eq:forcing_profile} with $k_0 = 18$. It oscillates in time, and its amplitude rises linearly over the first period, 
\begin{equation}
    G_k(t) = w \min(t / T, 1) \exp\!\Big( - \frac{\delta(k, k_0)^2}{2 \ell^2} \Big) \sin(\omega t),
\end{equation}
with $w = 10\,\sigma = 61.5$, $\omega = 2 \pi$ and the period $T = 2\pi / \omega = 1$, that is $100$ emulator steps. The run lasts $400$ steps: one period of rising amplitude followed by three periods at full amplitude. The emulators are those of the steady response experiment. The ensemble has $256$ independent members.

\paragraph{Capacity study.}
For each slow variable $k$, we compute the Wasserstein-1 distance $W_1(\mu_k, \mu_{\star,k})$ between the emulated marginal distribution and the true one, estimated from the long trajectory. The stationary error is
\begin{equation}
    \frac{1}{\sigma} \cdot \frac{1}{d} \sum_{k=0}^{d-1} W_1(\mu_k, \mu_{\star,k}).
\end{equation}
The response errors use the forcing \eqref{eq:forcing_profile} with amplitude $w = 10\,\sigma = 61.5$. The errors are averaged over $5$ independent ensembles of $2{,}048$ members. Each member is run twice from the same initial state and with the same random numbers, once without forcing and once with forcing. The mean response is the ensemble mean of the difference between the forced and the unforced states. The variance response is the ensemble variance of the forced states minus that of the unforced states. For each response, the error is the $L^2$ norm of the difference between the emulated and the true response over the $d$ sites and the first $50$ steps, divided by the norm of the true response. Both response errors are infinite when the stationary error is infinite.

\section{Additional experimental results} \label{section:additional-experimental-results}

\subsection{Oscillating forcing}

\begin{figure}[H]
    \centering
    \includegraphics[height=5cm]{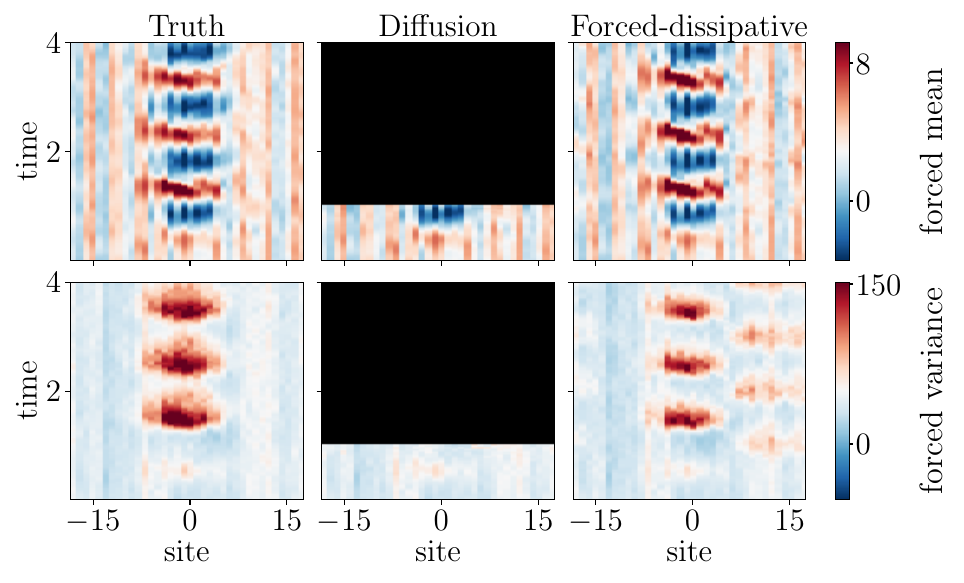}
    \caption{Ensemble mean and variance of the slow variables over all sites and times, under the oscillating forcing.}
    \label{figure:oscillatory_response_maps}
\end{figure}

The response maps for the oscillatory forcing experiment are provided in Figure \ref{figure:oscillatory_response_maps}.

\section{Additional mathematical results} \label{section:additional-mathematical-results}

\begin{proposition}[Explicit bound under time-varying forcing]
    \label{proposition:forced_bounded_explicit}
    Assume the drift structure \eqref{eq:dissipative_drift}--\eqref{eq:structural_properties} and Assumptions \ref{assumption:bounded_forcing}--\ref{assumption:regularity}. Let $u = (u_t)_{t \geq 0}$ be an arbitrary sequence of vectors of $\R^d$, not necessarily bounded, let $(X^u_t)$ be the perturbed chain of Definition \ref{definition:perturbations}, and let $\nu_t$ denote the law of $X^u_t$, with $\int V \ud \nu_0 < \infty$. Set
    \begin{equation}
        \label{eq:forcing_weighted_energy}
        a' := \frac{1 + e^{-\gamma}}{2} \in (0,1)
        \qquad \text{and} \qquad
        \bar{u}_t^{\,2} := (1 - a') \sum_{k=0}^{t-1} (a')^{\,t-1-k} \, \Vert u_k \Vert^2 ,
    \end{equation}
    the second quantity being the exponentially weighted average of the energy of the forcing up to time $t$. Then, for every $t \in \N$,
    \begin{equation}
        \label{eq:forced_moment_weighted}
        \int V \ud \nu_t
        \;\leq\;
        (a')^{\,t} \int V \ud \nu_0
        \;+\;
        \frac{2\,(e^{\gamma}+1)}{(e^{\gamma}-1)^2} \, \bar{u}_t^{\,2}
        \;+\;
        2 \Big( 1 + \frac{\Vert F \Vert_\infty^2}{\gamma^2} + \frac{d \, \sigma_{\max}^2}{\gamma} \Big) .
    \end{equation}
    If the forcing has amplitude $\sup_t \Vert u_t \Vert \leq U$, then $\bar{u}_t^{\,2} \leq U^2$ for every $t$, so that the right-hand side of \eqref{eq:forced_moment_weighted} is bounded uniformly in time, which proves Proposition \ref{proposition:forced_bounded}. If Assumption \ref{assumption:nonvanishing_noise} also holds, starting the perturbed chain at stationarity, $\nu_0 = \mu$, gives, for every observable $|\varphi| \leq V$,
    \begin{equation}
        \label{eq:forced_statistics_explicit}
        \sup_{t \in \N} \, \Big| \E \big[ \varphi(X^u_t) \big] \Big|
        \;\leq\;
        3 \Big( 1 + \frac{\Vert F \Vert_\infty^2}{\gamma^2} + \frac{d \, \sigma_{\max}^2}{\gamma} \Big)
        + \frac{2\,(e^{\gamma}+1)}{(e^{\gamma}-1)^2} \, U^2 .
    \end{equation}
\end{proposition}

\section{Proofs} \label{section:proofs}

\input{proofs.tex}

%% file: proofs.tex
\numberwithin{equation}{subsection}

\subsection{Well-posedness and regularity of the emulated kernel}
\label{section:proofs:wellposed}

In this section, we study the well-posedness and the regularity of the Markov kernel defined by the stochastic interpolant.

\begin{lemma}[Well-posedness]\label{lemma:wellposed}
    Under Assumptions \ref{assumption:bounded_forcing} and \ref{assumption:regularity}, for every $x_0 \in \R^d$ the stochastic differential equation \eqref{eq:generative_model_process} admits a unique strong, non-exploding solution $(X_s)_{s \in [0,1]}$, which satisfies
    \begin{equation}
        \label{eq:second_moment_bound}
        \sup_{s \in [0,1]} \E[\Vert X_s \Vert^2]
        \leq
        (1 + \Vert x_0 \Vert^2) \, e^{\kappa},
        \qquad
        \kappa := 1 + 2K + \Vert F \Vert_\infty^2 + d \sigma_{\max}^2 \in (0, \infty) .
    \end{equation}
    In particular, the transition kernel $P$ of the emulated chain \eqref{eq:emulated_trajectory} is well defined: $P(x, \cdot)$ is the law of $X_1$ in \eqref{eq:generative_model_process} started at $x_0 = x$.
\end{lemma}

\begin{proof}
    Under Assumptions \ref{assumption:bounded_forcing} and \ref{assumption:regularity}, the drift $b$ is continuous and of linear growth:
    \begin{equation}
        \label{eq:drift_linear_growth}
        \Vert b(x, s; x_0) \Vert \leq \Vert F \Vert_\infty + K \Vert x \Vert
        \qquad \text{for all } x, x_0 \in \R^d, \; s \in [0,1] .
    \end{equation}
    By Assumption \ref{assumption:regularity}, for every $r > 0$ there is  $K_r < \infty$ such that each of $F, A, D, S$ is $K_r$-Lipschitz in $(x, x_0)$ on the set $\{ \Vert x \Vert \leq r, \, \Vert x_0 \Vert \leq r \}$, for every $s \in [0,1]$. The drift is then itself locally Lipschitz in $(x, x_0)$ uniformly in $s$: for every $r > 0$, all $s \in [0,1]$ and all $x, x', x_0, x_0' \in \R^d$ of norm at most $r$,
    \begin{equation}
        \label{eq:drift_lipschitz}
        \Vert b(x, s; x_0) - b(x', s; x_0') \Vert
        \leq
        K_r' \big( \Vert x - x' \Vert + \Vert x_0 - x_0' \Vert \big),
        \qquad
        K_r' := (1 + 3r) K_r + K .
    \end{equation}
    For fixed $x_0$, the drift $(x, s) \mapsto b(x, s; x_0)$ and the noise coefficient $s \mapsto \sigma_s$ are continuous, the drift is Lipschitz in $x$ on every ball uniformly in $s$ by \eqref{eq:drift_lipschitz}, and both coefficients have linear growth by \eqref{eq:drift_linear_growth}. Let $\mathcal{L}_s$ denote the generator of \eqref{eq:generative_model_process} at time $s$, acting on twice continuously differentiable functions. For the function $V(x) = 1 + \Vert x \Vert^2$,
    \begin{equation}
        \mathcal{L}_s V(x)
        =
        2 \langle x, b(x, s; x_0) \rangle + d \sigma_s^2
        \qquad \text{for all } x \in \R^d, \; s \in [0,1] ,
    \end{equation}
    and the linear growth bound \eqref{eq:drift_linear_growth} together with $2 \Vert x \Vert \Vert F \Vert_\infty \leq \Vert x \Vert^2 + \Vert F \Vert_\infty^2$ gives
    \begin{equation}
        \mathcal{L}_s V(x)
        \leq
        2 \Vert x \Vert \Vert F \Vert_\infty + 2K \Vert x \Vert^2 + d \sigma_{\max}^2
        \leq
        (1 + 2K) \Vert x \Vert^2 + \Vert F \Vert_\infty^2 + d \sigma_{\max}^2
        \leq
        \kappa V(x) .
    \end{equation}
    Since moreover $V(x) \to \infty$ as $\Vert x \Vert \to \infty$, Theorem 3.5 of Chapter 3 in \citet{khasminskii2012stochastic} applies: the stochastic differential equation admits a non-exploding solution on $[0,1]$, unique up to indistinguishability, which satisfies $\E[V(X_s)] \leq V(x_0) e^{\kappa s}$ for all $s \in [0,1]$; this is \eqref{eq:second_moment_bound}. Pathwise uniqueness also follows from the local Lipschitz condition \eqref{eq:drift_lipschitz} alone \citep[Theorem 5.2.5]{karatzas1991brownian}, and together with the existence of a solution it implies that the solution is strong \citep[Corollary 5.3.23]{karatzas1991brownian}. See also \citet[Chapter 2]{mao2007stochastic}.
\end{proof}

All the solutions of \eqref{eq:generative_model_process} considered below are defined on the same filtered probability space $(\Omega, \mathcal{F}, (\mathcal{F}_s)_{s \in [0,1]}, \mathbb{P})$ and are driven by the same Brownian motion $W$, as allowed by Lemma \ref{lemma:wellposed}. We write $\omega$ for a generic outcome, that is a generic element of the sample space $\Omega$.

The kernel $P$ of Lemma \ref{lemma:wellposed} is defined pointwise in $x$. Two further properties are needed. Measurability of $x \mapsto P(x, A)$ makes $P$ a Markov kernel, so that the emulated trajectory \eqref{eq:emulated_trajectory} is a Markov chain. The Feller property, defined below, is the regularity that the stability analysis of Section \ref{section:proofs:ergodicity} relies on. Both follow from the same pathwise estimate, which we isolate first.

\begin{lemma}[Continuity in the initial condition]\label{lemma:continuity_initial_condition}
    Under Assumptions \ref{assumption:bounded_forcing} and \ref{assumption:regularity}, let $x \in \R^d$, let $(x_n)_{n \in \N}$ be a sequence of points of $\R^d$ converging to $x$, and let $X$ and $X^n$ solve \eqref{eq:generative_model_process} with initial conditions $X_0 = x$ and $X^n_0 = x_n$. Then
    \begin{equation}
        \label{eq:coupling_convergence}
        \sup_{s \in [0,1]} \Vert X^n_s - X_s \Vert \longrightarrow 0
        \qquad \text{almost surely as } n \to \infty .
    \end{equation}
\end{lemma}

\begin{proof}
    We couple the processes $X$ and $X^n$ by driving them with the \emph{same} Brownian motion $W$ and identical noise coefficients $\sigma_s$; both are given by Lemma \ref{lemma:wellposed}. Let $e^n_s := X^n_s - X_s \in \R^d$ for $s \in [0,1]$. Because the noise is shared, the diffusion terms cancel in the difference, yielding the integral equation
    \begin{equation}
        \label{eq:coupling_difference}
        e^n_s = (x_n - x) + \int_0^s \big[ b(X^n_u, u; x_n) - b(X_u, u; x) \big] \ud u
        \qquad \text{for all } s \in [0,1] .
    \end{equation}
    Each of the countably many equations \eqref{eq:generative_model_process}, the one for $X$ and those for the $X^n$ with $n \in \N$, holds outside a $\mathbb{P}$-null set, and each of the corresponding paths is continuous outside a $\mathbb{P}$-null set. Let $\Omega_0 \in \mathcal{F}$ be the complement of the union of these countably many null sets, so that $\mathbb{P}(\Omega_0) = 1$, and fix an outcome $\omega \in \Omega_0$ for the rest of the proof. Define
    \begin{equation}
        r := 2 + \Vert x \Vert + \sup_{s \in [0,1]} \Vert X_s(\omega) \Vert ,
    \end{equation}
    which is finite since $X$ has continuous paths. Let $K_r' = (1 + 3r) K_r + K$ be the Lipschitz constant of $b$ on the set $\{ \Vert x \Vert \leq r, \Vert x_0 \Vert \leq r \}$ given by \eqref{eq:drift_lipschitz}, and let
    \begin{equation}
        \tau_n := \inf \{ s \in [0,1] : \Vert e^n_s(\omega) \Vert \geq 1 \} \wedge 1 \in [0,1]
    \end{equation}
    be the first time at which the two paths are at distance $1$, or $1$ if this never happens. For $n$ large enough that $\Vert x_n - x \Vert \leq 1$, the four points $x$, $x_n$, $X_s(\omega)$ and $X^n_s(\omega)$ have norm at most $r$ for all $s \leq \tau_n$, since $\Vert X^n_s(\omega) \Vert \leq \Vert X_s(\omega) \Vert + \Vert e^n_s(\omega) \Vert \leq \Vert X_s(\omega) \Vert + 1$ on $[0, \tau_n]$. The Lipschitz bound \eqref{eq:drift_lipschitz} applied in \eqref{eq:coupling_difference} therefore gives, for $s \leq \tau_n$,
    \begin{equation}
        \Vert e^n_s(\omega) \Vert
        \leq
        \Vert x_n - x \Vert + K_r' \int_0^s \big( \Vert e^n_u(\omega) \Vert + \Vert x_n - x \Vert \big) \ud u
        \leq
        (1 + K_r') \Vert x_n - x \Vert + K_r' \int_0^s \Vert e^n_u(\omega) \Vert \ud u ,
    \end{equation}
    and Gr\"onwall's inequality yields the pathwise bound
    \begin{equation}
        \sup_{s \in [0, \tau_n]} \Vert e^n_s(\omega) \Vert
        \leq
        (1 + K_r') \, e^{K_r'} \, \Vert x_n - x \Vert =: \epsilon_n .
    \end{equation}
    Note that $\epsilon_n \geq 0$ tend to $0$ as $n \to \infty$. As soon as $\epsilon_n < 1$, the distance $1$ is never reached: otherwise $\Vert e^n_{\tau_n}(\omega) \Vert = 1$ by continuity, contradicting the bound. Hence $\tau_n = 1$ and $\sup_{s \in [0,1]} \Vert e^n_s(\omega) \Vert \leq \epsilon_n \to 0$. This holds for every $\omega \in \Omega_0$, which proves \eqref{eq:coupling_convergence}.
\end{proof}

\begin{proposition}[Markov kernel]\label{proposition:markov_kernel}
    Under Assumptions \ref{assumption:bounded_forcing} and \ref{assumption:regularity}, the map $x \mapsto P(x, A)$ is Borel measurable for every $A \in \mathcal{B}(\R^d)$. Together with the fact that $P(x, \cdot)$ is a probability measure for every $x$, this makes $P$ a Markov kernel on $\R^d$, and \eqref{eq:emulated_trajectory} a Markov chain with transition kernel $P$.
\end{proposition}

\begin{proof}
    For a bounded Borel function $f : \R^d \to \R$, write $Pf(x) := \E[f(X_1) \mid X_0 = x]$, so that $|Pf(x)| \leq \Vert f \Vert_\infty$ for all $x \in \R^d$. We first prove that $Pf$ is continuous when $f$ is bounded and continuous. Fix $x \in \R^d$ and a sequence $(x_n)_{n \in \N}$ of points of $\R^d$ converging to $x$, and let $X$ and $X^n$ be as in Lemma \ref{lemma:continuity_initial_condition}. That lemma gives $X^n_1 \to X_1$ almost surely, and $f$ is continuous, so $f(X^n_1) \to f(X_1)$ almost surely. Because $f$ is bounded, the bounded convergence theorem yields
    \begin{equation}
        \label{eq:pf_continuity}
        Pf(x_n) = \E[f(X^n_1)] \longrightarrow \E[f(X_1)] = Pf(x) \quad \text{as } n \to \infty .
    \end{equation}
    Since $x$ and the sequence $(x_n)$ were arbitrary, $Pf$ is continuous everywhere, hence Borel measurable.

    Let now $\mathcal{H}$ be the class of bounded Borel functions $f : \R^d \to \R$ such that $Pf$ is Borel measurable. It is a vector space containing the constant functions, it is closed under bounded monotone limits by the dominated convergence theorem, and it contains the bounded continuous functions by the previous paragraph. The bounded continuous functions are stable under pointwise multiplication and generate $\mathcal{B}(\R^d)$, so the monotone class theorem applies: $\mathcal{H}$ contains all bounded Borel functions, in particular the indicator functions $\mathbbm{1}_A$ for $A \in \mathcal{B}(\R^d)$. Since $P \mathbbm{1}_A = P(\cdot, A)$, the map $x \mapsto P(x, A)$ is Borel measurable.
\end{proof}

\begin{definition}[Feller kernel]\label{definition:feller}
    A Markov kernel $Q$ on $\R^d$ is Feller if $Qf : x \mapsto \int f(y) \, Q(x, \ud y)$ is bounded and continuous whenever $f : \R^d \to \R$ is bounded and continuous.
\end{definition}

The Feller property is the stochastic counterpart of continuity of the dynamics: nearby initial states produce nearby one-step statistics. It is the regularity that lets pointwise properties of the kernel extend uniformly to compact sets, and the emulated chain inherits it from the continuous dependence of the solution of \eqref{eq:generative_model_process} on its initial condition.

\begin{lemma}[Feller property]\label{lemma:feller}
    Under Assumptions \ref{assumption:bounded_forcing} and \ref{assumption:regularity}, the Markov kernel $P$ is Feller.
\end{lemma}

\begin{proof}
    Let $f : \R^d \to \R$ be bounded and continuous. Boundedness of $Pf$ is immediate, since $|Pf(x)| \leq \Vert f \Vert_\infty$ for all $x \in \R^d$. Continuity of $Pf$ is \eqref{eq:pf_continuity}, established in the first paragraph of the proof of Proposition \ref{proposition:markov_kernel} from the pathwise convergence of Lemma \ref{lemma:continuity_initial_condition}.
\end{proof}

\subsection{Proof of Proposition \ref{proposition:energy_bound}}

\energybound*

\begin{proof}
    Fix $x_0 \in \R^d$ as the initial state $X_0 = x_0$, let $X$ be the solution given by Lemma \ref{lemma:wellposed}, and write $E(s) := \E[\|X_s\|^2 \mid X_0 = x_0]$ for the expected energy, which is bounded on $[0,1]$ by \eqref{eq:second_moment_bound}. To establish the Foster--Lyapunov drift condition of Proposition \ref{proposition:foster}, we seek a uniform bound on $E(1) = \E[\|X_1\|^2 \mid X_0 = x_0]$.

    Applying It\^o's formula to $\|X_s\|^2$ along the stochastic interpolant dynamics $\ud X_s = b(X_s, s; x_0) \ud s + \sigma_s \ud W_s$ yields
    \begin{equation}
        \ud \|X_s\|^2 = 2\langle X_s, \ud X_s \rangle + \langle \ud X_s, \ud X_s \rangle = 2\langle X_s, b(X_s, s; x_0) \rangle \ud s + 2\langle X_s, \sigma_s \ud W_s \rangle + d\sigma_s^2 \ud s ,
    \end{equation}
    that is, in integrated form,
    \begin{equation}
        \label{eq:ito_energy}
        \|X_s\|^2
        =
        \|x_0\|^2
        + \int_0^s \big( 2 \langle X_r, b(X_r, r; x_0) \rangle + d \sigma_r^2 \big) \ud r
        + 2 \int_0^s \sigma_r \langle X_r, \ud W_r \rangle
        \qquad \text{for all } s \in [0,1] .
    \end{equation}
    The stochastic integral in \eqref{eq:ito_energy} is a square-integrable martingale, since
    \begin{equation}
        \E \int_0^1 \sigma_r^2 \|X_r\|^2 \ud r \leq \sigma_{\max}^2 \sup_{r \in [0,1]} E(r) < \infty ,
    \end{equation}
    so its expectation vanishes. The integrand of the Lebesgue integral is dominated, by the linear growth \eqref{eq:drift_linear_growth}, as
    \begin{equation}
        \big| 2 \langle X_r, b(X_r, r; x_0) \rangle + d \sigma_r^2 \big|
        \leq
        2 \|X_r\| \big( \|F\|_\infty + K \|X_r\| \big) + d \sigma_{\max}^2 ,
    \end{equation}
    which is integrable on $[0,1] \times \Omega$ by \eqref{eq:second_moment_bound}. Taking the expectation in \eqref{eq:ito_energy} and applying Fubini's theorem thus yields the evolution of the second moment,
    \begin{equation}
        \label{eq:energy_integrated}
        E(s) = \|x_0\|^2 + \int_0^s \big( 2 \, \E[\langle X_r, b(X_r, r; x_0) \rangle] + d \sigma_r^2 \big) \ud r
        \qquad \text{for all } s \in [0,1] ,
    \end{equation}
    an absolutely continuous function of $s$, so that
    \begin{equation}
        \label{eq:energy_derivative}
        \frac{\ud}{\ud s} \E[\|X_s\|^2] = 2 \, \E[\langle X_s, b(X_s, s; x_0) \rangle] + d \sigma_s^2
        \qquad \text{for almost every } s \in [0,1] .
    \end{equation}
    Substituting the drift decomposition $b(X_s, s; x_0) = F + A X_s + D X_s + S X_s$, where the four components are evaluated at $z = (X_s, s; x_0)$, expands the drift term into four components:
    \begin{align*}
        2 \E[\langle X_s, b(X_s, s; x_0) \rangle] = \; & 2 \E[\langle X_s, F \rangle] + 2 \E[\langle X_s, A X_s \rangle]         \\
                                                       & + 2 \E[\langle X_s, D X_s \rangle] + 2 \E[\langle X_s, S X_s \rangle] .
    \end{align*}

    We now bound each of the four terms individually, pointwise in $z$, using \eqref{eq:structural_properties}:
    \begin{enumerate}
        \item \textbf{Forcing term:} By Cauchy--Schwarz and Young's inequality ($2ab \le \epsilon a^2 + \epsilon^{-1} b^2$, with $\epsilon = \gamma$) and Assumption \ref{assumption:bounded_forcing},
              \begin{equation}
                  2 \langle X_s, F \rangle \le 2 \|X_s\| \|F\| \le \gamma \|X_s\|^2 + \frac{\|F\|_\infty^2}{\gamma} .
              \end{equation}
        \item \textbf{Linear conservative term:} By the skew-symmetry of $A$ ($A^\top = -A$),
              \begin{equation}
                  \langle X_s, A X_s \rangle = X_s^\top A X_s = 0 .
              \end{equation}
        \item \textbf{Dissipative term:} By the dissipativity constraint ($D \preceq -\gamma I_d$),
              \begin{equation}
                  2 \langle X_s, D X_s \rangle \le -2 \gamma \|X_s\|^2 .
              \end{equation}
        \item \textbf{Nonlinear term:} By the constraint $\langle y, S y \rangle \leq 0$ for all $y \in \R^d$,
              \begin{equation}
                  \langle X_s, S X_s \rangle \le 0 .
              \end{equation}
    \end{enumerate}

    Substituting these bounds back into \eqref{eq:energy_derivative} gives, for almost every $s \in [0,1]$, the differential inequality
    \begin{equation}
        \frac{\ud}{\ud s} \E[\|X_s\|^2] \le - \gamma \, \E[\|X_s\|^2] + \frac{\|F\|_\infty^2}{\gamma} + d \sigma_s^2 ,
    \end{equation}
    which is \eqref{eq:expected_energy_bound} after division by $2$.
\end{proof}

\subsection{Proof of Proposition \ref{proposition:foster}}

\fosterlyapunov*

\begin{proof}
    Condition on $X_t = x$, that is, run \eqref{eq:generative_model_process} from $x_0 = x$, and let $E : [0,1] \to \R_+$, $E(s) := \E[\Vert X_s \Vert^2]$, be the expected energy, which is absolutely continuous by Proposition \ref{proposition:energy_bound}. Define
    \begin{equation}
        g := \frac{\Vert F \Vert_\infty^2}{\gamma} + d \sigma_{\max}^2 \in [0, \infty) .
    \end{equation}
    Proposition \ref{proposition:energy_bound} and the bound $\sigma_s \leq \sigma_{\max}$ give
    \begin{equation}
        E'(s) \leq - \gamma \, E(s) + g
        \qquad \text{for almost every } s \in [0,1] .
    \end{equation}
    We apply Gr\"onwall's inequality in its differential form, via the integrating factor $e^{\gamma s}$. The function $s \mapsto e^{\gamma s} E(s)$ is absolutely continuous on $[0,1]$, as a product of absolutely continuous functions, and its derivative satisfies
    \begin{equation}
        \frac{\ud}{\ud s} \big( e^{\gamma s} E(s) \big)
        =
        e^{\gamma s} \big( E'(s) + \gamma E(s) \big)
        \leq
        e^{\gamma s} g
        \qquad \text{for almost every } s \in [0,1] .
    \end{equation}
    Integrating over $[0,1]$ yields
    \begin{equation}
        e^{\gamma} E(1) - E(0)
        \leq
        g \int_0^1 e^{\gamma s} \ud s
        =
        g \, \frac{e^{\gamma} - 1}{\gamma} ,
    \end{equation}
    that is, since $E(0) = \Vert x \Vert^2$,
    \begin{equation}
        \E[\Vert X_1 \Vert^2 \mid X_0 = x]
        =
        E(1)
        \leq
        e^{-\gamma} \Vert x \Vert^2 + \frac{1 - e^{-\gamma}}{\gamma} \left( \frac{\Vert F \Vert_\infty^2}{\gamma} + d \sigma_{\max}^2 \right) .
    \end{equation}

    Recognizing that $X_1$ is exactly the next step of the Markov chain (i.e., $X_{t+1}$ conditioned on $X_t = x$), and adding $1 = e^{-\gamma} + (1 - e^{-\gamma})$ to both sides, this translates to the Foster--Lyapunov drift condition
    \begin{equation}
        PV(x) = \E[V(X_{t+1}) \mid X_t = x] \le a V(x) + c ,
    \end{equation}
    with geometric contraction factor $a := e^{-\gamma} \in (0, 1)$ and constant
    \begin{equation}
        c := (1 - e^{-\gamma}) \left( 1 + \frac{\Vert F \Vert_\infty^2}{\gamma^2} + \frac{d \sigma_{\max}^2}{\gamma} \right) \in (0, \infty) .
    \end{equation}
\end{proof}

\subsection{Proof of Proposition \ref{proposition:ergodicity}}
\label{section:proofs:ergodicity}

\ergodicity*

The proof also gives an explicit bound on the second moment of the invariant measure,
\begin{equation}
    \label{eq:invariant_moment_bound}
    \int V \ud \mu
    \leq
    \frac{c}{1-a}
    =
    1 + \frac{\Vert F \Vert_\infty^2}{\gamma^2} + \frac{d \, \sigma_{\max}^2}{\gamma},
\end{equation}
with $a$ and $c$ as in Proposition \ref{proposition:foster}. This bound is used in Appendix \ref{section:additional-mathematical-results}.

The proof relies on verifying the conditions of the geometric ergodicity theorem for general state space Markov chains \citep[Theorem 15.0.1]{meyn2012markov}. Three arguments are required:
\begin{enumerate}
    \item irreducibility and aperiodicity, which we define below and establish in Lemma \ref{lemma:irreducible_aperiodic},
    \item {the petite set property}, established in Lemma \ref{lemma:petite},
    \item a geometric drift condition, following from the Foster--Lyapunov drift condition \eqref{eq:Foster_Lyapunov} established above.
\end{enumerate}

We now define the relevant mathematical objects.

\begin{definition}[Irreducibility]\label{definition:irreducibility}
    A Markov kernel $Q$ on $\R^d$ is {$\mathrm{Leb}$-irreducible} if for every $x \in \R^d$ and every $A \in \mathcal{B}(\R^d)$ with $\mathrm{Leb}(A) > 0$ there exists $t \in \N$ such that $Q^t(x, A) > 0$.
\end{definition}

Irreducibility states that no region of positive Lebesgue measure is unreachable: the state space cannot split into disjoint basins, each carrying its own equilibrium, so the chain admits at most one statistical equilibrium.

\begin{definition}[Aperiodicity]\label{definition:aperiodicity}
    A $\mathrm{Leb}$-irreducible kernel $Q$ admits a cyclic decomposition: disjoint sets $D_0, \dots, D_{p-1} \in \mathcal{B}(\R^d)$ with $Q(x, D_{i+1 \bmod p}) = 1$ for $x \in D_i$, covering $\R^d$ up to a $\mathrm{Leb}$-null set \citep[Theorem 5.4.4]{meyn2012markov}. The largest such integer $p \geq 1$ is the \emph{period} of the chain, which is \emph{aperiodic} if $p = 1$.
\end{definition}

Aperiodicity rules out a hidden clock: a chain of period $p \geq 2$ rotates through the classes $D_0, \dots, D_{p-1}$ deterministically, and its marginal law oscillates forever instead of converging, even when its time averages converge.

\begin{definition}[Petite set]\label{definition:petite}
    A set $C \in \mathcal{B}(\R^d)$ is petite for the kernel $Q$ if there exist a probability distribution $(\eta_t)_{t \in \N}$ on $\N$ and a nontrivial measure $\nu$ on $\mathcal{B}(\R^d)$ such that
    \begin{equation}
        \sum_{t \in \N} \eta_t \, Q^t(x, A) \geq \nu(A)
        \quad \text{for all} \; x \in C, \; A \in \mathcal{B}(\R^d).
    \end{equation}
    When $(\eta_t)$ is concentrated at a single time, $C$ is called a small set \citep[Chapter 5]{meyn2012markov}.
\end{definition}

A petite set is one on which the chain regenerates uniformly: started from any $x \in C$ and sampled at an independent randomized time, the chain deposits at least the common measure $\nu$, from which two copies started anywhere in $C$ can be coupled with positive probability. Petite sets thus play, in a general state space, the role that individual states play for countable chains.

The first two arguments only depend on two properties of the transition kernel: the Feller property of Lemma \ref{lemma:feller}, and the everywhere positivity of Proposition \ref{proposition:positivity} below. We therefore state them for a generic Markov kernel $Q$ on $\R^d$, so that they can be reused for the perturbed kernels $P_v$ in Section \ref{section:proofs:forced}.

\begin{lemma}[Irreducibility and aperiodicity]
    \label{lemma:irreducible_aperiodic}
    Let $Q$ be a Markov kernel on $\R^d$ such that $Q(x, A) > 0$ for every $x \in \R^d$ and every $A \in \mathcal{B}(\R^d)$ with $\mathrm{Leb}(A) > 0$. Then $Q$ is irreducible for the Lebesgue measure and aperiodic.
\end{lemma}

\begin{lemma}[Compact sets are petite]
    \label{lemma:petite}
    Let $Q$ be a Feller Markov kernel on $\R^d$ such that $Q(x, A) > 0$ for every $x \in \R^d$ and every $A \in \mathcal{B}(\R^d)$ with $\mathrm{Leb}(A) > 0$. Then every compact set of $\R^d$ is petite for $Q$.
\end{lemma}

\begin{proof}[Proof of Lemma \ref{lemma:irreducible_aperiodic}]
    Irreducibility is immediate from the positivity assumption on $Q$, with $t = 1$ in Definition \ref{definition:irreducibility}. For aperiodicity, suppose the period is $p \geq 2$, and let $D_0, \dots, D_{p-1}$ be the cyclic classes of Definition \ref{definition:aperiodicity}. Since the classes cover $\R^d$ up to a $\mathrm{Leb}$-null set, $\mathrm{Leb}(D_j) > 0$ for some $j$; in particular $D_j$ is nonempty. For $x \in D_j$, the decomposition gives $Q(x, D_{j+1 \bmod p}) = 1$, and $D_{j+1 \bmod p}$ is disjoint from $D_j$ since $p \geq 2$, so $Q(x, D_j) = 0$. This contradicts the positivity assumption. Hence $p = 1$.
\end{proof}

\begin{proof}[Proof of Lemma \ref{lemma:petite}]
    The kernel $Q$ is Feller by assumption. Equivalently, $Q(\cdot, O)$ is lower semicontinuous for every open set $O$, which is the form of the Feller property used in \citet[Chapter 6, Proposition 6.1.1]{meyn2012markov}. By Lemma \ref{lemma:irreducible_aperiodic}, the chain is irreducible for the Lebesgue measure, so its maximal irreducibility measure $\psi$ dominates $\mathrm{Leb}$ \citep[Proposition 4.2.2]{meyn2012markov}. In particular $\operatorname{supp} \psi = \R^d$ has nonempty interior. By \cite[Theorem 6.2.9]{meyn2012markov}, the chain is a T-chain. By \cite[Theorem 6.0.1]{meyn2012markov}, for a $\psi$-irreducible T-chain, every compact set is petite.
\end{proof}

We showed that $P$ is Feller, and the Foster--Lyapunov condition in the previous sections. Irreducibility of $P$ is shown with the following result, whose proof is deferred to the next section.

\begin{proposition}[Everywhere positivity]
    \label{proposition:positivity}
    Under Assumptions \ref{assumption:bounded_forcing}--\ref{assumption:nonvanishing_noise}, $P(x, A) > 0$ for every $x \in \R^d$ and every $A \in \mathcal{B}(\R^d)$ with $\mathrm{Leb}(A) > 0$.
\end{proposition}

Given these results, we now prove geometric ergodicity of $P$.

\begin{proof}[Proof of Proposition \ref{proposition:ergodicity}]
    We establish the three arguments mentioned above for the emulated kernel $P$.

    We start by irreducibility, aperiodicity and petite sets. By Proposition \ref{proposition:positivity}, $P(x, A) > 0$ for every $x \in \R^d$ and every $A \in \mathcal{B}(\R^d)$ with $\mathrm{Leb}(A) > 0$, and by Lemma \ref{lemma:feller} the kernel $P$ is Feller. Lemmas \ref{lemma:irreducible_aperiodic} and \ref{lemma:petite} therefore apply to $Q = P$: the emulated chain is irreducible for the Lebesgue measure and aperiodic, and its compact sets are petite. These are the first two arguments.

    We now show the geometric drift condition. For $x \in \R^d$, let
    \begin{equation}
        \Delta V (x)
        :=
        P V(x) - V(x)
    \end{equation}
    be the expected Lyapunov increment, or Lyapunov drift; this defines a function $\Delta V : \R^d \to \R$.
    The Foster--Lyapunov condition \eqref{eq:Foster_Lyapunov} of Proposition \ref{proposition:foster}, which is the third argument, reads
    \begin{equation}
        \Delta V(x)
        \leq
        -(1-a) V(x) + c
        \quad \text{for all} \; x \in \R^d,
    \end{equation}
    which we reformulate as
    \begin{equation}
        \label{eq:drift_split}
        \Delta V(x)
        \leq
        -\lambda V(x) -  \frac{1-a}{2}V(x)+ c
    \end{equation}
    with
    \begin{equation}
        \lambda:=(1-a)/2 >0.
    \end{equation}
    Fix $R > 1 + 2c/(1-a)$ and define the level set
    \begin{equation}
        C := \{V \leq R\} = \{ x \in \R^d \mid \Vert x \Vert^2 \leq R-1\} ,
    \end{equation}
    a closed ball, nonempty since $R > 1$, hence a compact subset of $\R^d$. The threshold $R$ is chosen large enough such that
    \begin{equation}
        x \notin C \implies -\frac{1-a}{2} V(x) + c \leq 0.
    \end{equation}
    For $x\in C$, the same term is bounded by $c$.
    Then, \eqref{eq:drift_split} implies the geometric drift condition
    \begin{equation}
        \Delta V(x)
        \leq
        -\lambda V(x) + c \, \mathbbm{1}_C(x)
        \quad \text{for all} \; x \in \R^d,
    \end{equation}
    of \citet[Theorem 15.0.1]{meyn2012markov}. The set $C$ is compact, hence petite by Lemma \ref{lemma:petite}.

    We can now apply \citet[Theorem 15.0.1]{meyn2012markov}: the chain is positive recurrent, hence admits an invariant probability measure $\mu$, unique by irreducibility \citep[Theorem 10.0.1]{meyn2012markov}, and, since $V$ is finite on all of $\R^d$, there exist  $r > 1$ and $M < \infty$ such that
    \begin{equation}
        \sum_{t \in \N} r^t \, \big\Vert P^t(x, \cdot) - \mu \big\Vert_V \leq M V(x)
        \quad \text{for all} \; x \in \R^d ,
    \end{equation}
    where the $V$-norm of a signed measure $\nu$ is defined by
    \begin{equation}
        \Vert \nu \Vert_V := \sup_{|f| \leq V} \Big| \int f \ud \nu \Big| ,
    \end{equation}
    the supremum being over the measurable functions $f : \R^d \to \R$ with $|f| \leq V$.
    Bounding each term of this sum by the sum itself yields
    \begin{equation}
        \label{eq:ergodicity_V}
        \big\Vert P^t(x, \cdot) - \mu \big\Vert_V \leq M \rho^t V(x)
        \qquad \text{for all } x \in \R^d \text{ and } t \in \N ,
    \end{equation}
    with $\rho := 1/r \in (0,1)$.

    \emph{Moment bound.} Iterating the Foster--Lyapunov condition \eqref{eq:Foster_Lyapunov} gives
    \begin{equation}
        \label{eq:iterated_drift}
        P^t V(x) \leq a^{\,t} V(x) + c \sum_{k=0}^{t-1} a^{\,k} \leq a^{\,t} V(x) + \frac{c}{1-a}
        \qquad \text{for all } x \in \R^d \text{ and } t \in \N .
    \end{equation}
    For $\ell > 0$, the function $V \wedge \ell$ is bounded and measurable, so the total variation convergence implied by \eqref{eq:ergodicity_V} gives, for any $x \in \R^d$,
    \begin{equation}
        \int (V \wedge \ell) \ud \mu = \lim_{t \to \infty} P^t (V \wedge \ell)(x) \leq \liminf_{t \to \infty} P^t V(x) \leq \frac{c}{1-a} .
    \end{equation}
    This nondecreasing function of $\ell$ is bounded, hence its limit is finite. Monotone convergence as $\ell \to \infty$ yields
    \begin{equation}
        \int V \ud \mu \leq \frac{c}{1-a} = 1 + \frac{\Vert F \Vert_\infty^2}{\gamma^2} + \frac{d \, \sigma_{\max}^2}{\gamma} ,
    \end{equation}
    which is \eqref{eq:invariant_moment_bound}.

    \emph{Convergence in total variation.} Since $V \geq 1$, every measurable $f$ with $|f| \leq 1$ satisfies $|f| \leq V$, so that
    \begin{equation}
        \big\Vert P^t(x, \cdot) - \mu \big\Vert_{\mathrm{TV}}
        \leq
        \big\Vert P^t(x, \cdot) - \mu \big\Vert_V
        \leq
        M \rho^t V(x) \xrightarrow[t \to \infty]{} 0
        \qquad \text{for all } x \in \R^d ,
    \end{equation}
    which is the total variation bound \eqref{eq:geometric_ergodicity} together with the convergence \eqref{eq:stability_definition} required by Definition \ref{definition:stability}.
\end{proof}

\subsection{Proof of Proposition \ref{proposition:positivity}}

\begin{proof}
    Let $\tilde{X}_s := x_0 + \int_0^s \sigma_r \, \ud W_r$ be the driftless reference process driven by the same Brownian motion. By Assumption \ref{assumption:nonvanishing_noise}, $\sigma_s \ge \sigma_{\min} > 0$, so the covariance matrix $\Sigma := \big( \int_0^1 \sigma_s^2 \, \ud s \big) I_d$ is positive definite. Thus, $\tilde{X}_1 \sim \mathcal{N}(x_0, \Sigma)$ admits a strictly positive Gaussian density $\tilde{p}(y) > 0$ for all $y \in \R^d$, the It\^o integral of a deterministic square-integrable integrand against a Brownian motion being a centered Gaussian vector with covariance given by the It\^o isometry.

    To relate the law of $X_1$ to $\tilde{X}_1$, define $u_s := \sigma_s^{-1} b(X_s, s; x_0)$, a continuous adapted process since $\sigma_s \geq \sigma_{\min} > 0$, and the exponential local martingale
    \begin{equation}
        Z_s := \exp\left( -\int_0^s \langle u_r, \ud W_r \rangle - \frac{1}{2} \int_0^s \|u_r\|^2 \ud r \right).
    \end{equation}
    Assume for the moment that $Z$ is a true martingale on $[0,1]$. Defining $\mathbb{Q}$ via $\ud \mathbb{Q} / \ud \mathbb{P} := Z_1$ on $\mathcal{F}_1$, Girsanov's theorem \citep[Theorem 3.5.1]{karatzas1991brownian} yields that $W_s^\mathbb{Q} := W_s + \int_0^s u_r \ud r$ is a $\mathbb{Q}$-Brownian motion, and the It\^o integrals computed under the two measures are related by \citet[Problem 3.5.6]{karatzas1991brownian}, applied to the bounded deterministic integrand $\sigma$. Under $\mathbb{Q}$ the equation becomes $\ud X_s = b(X_s, s; x_0) \ud s + \sigma_s (\ud W_s^\mathbb{Q} - u_s \ud s) = \sigma_s \ud W_s^\mathbb{Q}$, since $\sigma_s u_s = b(X_s, s; x_0)$, so that $X_1 = x_0 + \int_0^1 \sigma_r \ud W^\mathbb{Q}_r$ has the same centered Gaussian form as $\tilde{X}_1$, that is $\mathrm{Law}_\mathbb{Q}(X_1) = \mathcal{N}(x_0, \Sigma)$. Since $Z_1 > 0$ $\mathbb{P}$-a.s., $\mathbb{P}$ and $\mathbb{Q}$ are mutually equivalent, yielding
    \begin{equation}
        \mathbb{P}(X_1 \in A) > 0 \iff \mathbb{Q}(X_1 \in A) = \int_A \tilde{p}(y) \ud y > 0 ,
    \end{equation}
    which is the claim.

    It remains to show that $Z$ is a true martingale. By the linear growth \eqref{eq:drift_linear_growth}, $\|b(X_s, s; x_0)\| \le \|F\|_\infty + K \|X_s\|$, and since $\sigma_s \ge \sigma_{\min}$, we have $\|u_s\| \le \sigma_{\min}^{-1}(\|F\|_\infty + K \|X_s\|)$. Squaring and using $(a+b)^2 \le 2a^2 + 2b^2$ gives $\|u_s\|^2 \le C_1 + C_2 \|X_s\|^2$ for the constants $C_1 := 2\|F\|_\infty^2 / \sigma_{\min}^2$ and $C_2 := 2K^2 / \sigma_{\min}^2$. We use  the partitioned form of the criterion \citep[Corollary 3.5.14]{karatzas1991brownian}, which asks for an increasing sequence of times $0 = t_0 < t_1 < \dots$ tending to infinity such that
    \begin{equation} \label{eq:patchcond}
        \E\left[\exp\left(\frac{1}{2}\int_{t_{k-1}}^{t_k} \|u_s\|^2 \ud s\right)\right] < \infty \qquad \text{for every } k \geq 1 ,
    \end{equation}
    together with a global bound on the supremum of $\Vert X_s \Vert$ to verify it. We take $t_k := k \delta$ with $\delta := 1/n$ for $k \le n$, the integer $n$ being chosen at the end of the proof, and $t_k := k - n + 1$ for $k > n$, extending $u$ by $0$ after time $1$ so that the blocks beyond $1$ contribute the value $1$.

    To establish \eqref{eq:patchcond} we apply Gr\"onwall's inequality over the entire interval $[0,1]$, to bound $\Xi := \sup_{s \in [0,1]} \|X_s\|^2$. Integrating the SDE and using the linear growth of $b$ yields, for any $s \in [0,1]$,
    \begin{equation}
        \|X_s\| \le \|x_0\| + \|F\|_\infty + K\int_0^s \|X_r\| \ud r + B^* ,
        \qquad B^* := \sup_{s \in [0,1]} \Big\| \int_0^s \sigma_r \ud W_r \Big\| .
    \end{equation}
    By Gr\"onwall's inequality, $\sup_{s \in [0,1]} \|X_s\| \le (\|x_0\| + \|F\|_\infty + B^*) e^{K}$, and squaring with $(a+b+c)^2 \le 3(a^2+b^2+c^2)$ yields the global bound
    \begin{equation}
        \Xi \le 3 e^{2K} \left( \|x_0\|^2 + \|F\|_\infty^2 + (B^*)^2 \right).
    \end{equation}
    The process $B_s := \int_0^s \sigma_r \ud W_r$ is a continuous martingale whose coordinates have the deterministic quadratic variation $\langle B^i \rangle_s = \int_0^s \sigma_r^2 \ud r \le \sigma_{\max}^2$, so Novikov's criterion \citep[Corollary 3.5.13]{karatzas1991brownian} makes $\exp(\lambda B^i_s - \frac{\lambda^2}{2} \langle B^i \rangle_s)$ a positive martingale of expectation $1$ for every $\lambda > 0$. Doob's maximal inequality applied to it with $\lambda := a / \sigma_{\max}^2$ gives $\mathbb{P}(\sup_{s \le 1} B^i_s \ge a) \le e^{-a^2/(2\sigma_{\max}^2)}$ for every $a > 0$, and applying this to $\pm B^i$, the variable $R_i := \sup_{s \in [0,1]} |B^i_s|$ satisfies $\mathbb{P}(R_i \ge a) \le 2 e^{-a^2/(2\sigma_{\max}^2)}$. For any $\theta < \theta_0 := 1/(2\sigma_{\max}^2)$, integrating this tail bound gives
    \begin{equation}
        \E[\exp(\theta R_i^2)] = 1 + \int_0^\infty 2\theta a e^{\theta a^2} \, \mathbb{P}(R_i > a) \, \ud a \le 1 + \frac{4\theta\sigma_{\max}^2}{1 - 2\theta\sigma_{\max}^2} < \infty .
    \end{equation}
    Since the $d$ coordinates of $W$ are independent, so are the $R_i$, and $(B^*)^2 \le \sum_{i=1}^d R_i^2$ gives $\E[\exp(\theta (B^*)^2)] < \infty$ for all $\theta < \theta_0$.

    For any block $k \le n$, using $\|u_s\|^2 \le C_1 + C_2 \|X_s\|^2 \le C_1 + C_2 \Xi$ and the global bound on $\Xi$,
    \begin{equation}
        \frac{1}{2}\int_{t_{k-1}}^{t_k} \|u_s\|^2 \ud s
        \le
        \frac{\delta}{2} \Big( C_1 + 3 C_2 e^{2K} \big( \|x_0\|^2 + \|F\|_\infty^2 \big) \Big)
        + \frac{3 \delta C_2 e^{2K}}{2} (B^*)^2 .
    \end{equation}
    The first term is a finite deterministic constant for fixed $x_0$, and the second is $\theta (B^*)^2$ with $\theta := 3 \delta C_2 e^{2K} / 2$. Choosing $n \ge 3 C_2 e^{2K} \sigma_{\max}^2 + 1$ ensures $\theta < \theta_0$, which bounds \eqref{eq:patchcond} uniformly in $k$, proving that $Z$ is a true martingale and completing the proof.
\end{proof}

\subsection{Proofs of Propositions \ref{proposition:forced_bounded} and \ref{proposition:forced_bounded_explicit}}

\forcedstability*

\begin{lemma}
    \label{lemma:perturbation_energy}
    For every $v \in \mathbb{R}^d$ and every $\varepsilon > 0$,
    \begin{equation}
        \label{eq:perturbed_drift}
        P_v V(x) \leq a(1+\varepsilon) V(x)
        + a\left(1 + \frac{1}{\varepsilon}\right)\lVert v\rVert^2 + c
        \qquad \text{for all } x \in \R^d ,
    \end{equation}
    with $a$ and $c$ as in Proposition~\ref{proposition:foster}.
\end{lemma}

\begin{proof}
    The Foster--Lyapunov condition of Proposition~\ref{proposition:foster} states $PV(x) \leq aV(x) + c$ for all $x$. By
    definition $P_vV(x) = (PV)(x+v)$, so
    \begin{equation}
        \label{eq:perturbed_energy_inequality}
        P_vV(x)\leq aV(x+v) + c.
    \end{equation}
    Applying the Cauchy--Schwarz inequality and Young's inequality, for all $\varepsilon > 0$,
    \begin{equation}
        2\langle x, v\rangle \leq \varepsilon\lVert x\rVert^2 + \varepsilon^{-1}\lVert v\rVert^2.
    \end{equation}
    We then obtain
    \begin{equation}
        \begin{aligned}
            V(x+v) & = 1 + \lVert x\rVert^2 + 2\langle x,v\rangle + \lVert v\rVert^2
            \\
                   & \leq 1 + (1+\varepsilon)\lVert x\rVert^2 + \Big(1+\frac1\varepsilon\Big)\lVert v\rVert^2 .
        \end{aligned}
    \end{equation}
    Finally, we bound
    \begin{equation}
        (1+\varepsilon)V(x) = (1+\varepsilon) + (1+\varepsilon)\lVert x\rVert^2
        \geq 1 + (1+\varepsilon)\lVert x\rVert^2
    \end{equation}
    to obtain
    \begin{equation}
        V(x+v)  \leq (1+\varepsilon)V(x) + \Big(1+\frac1\varepsilon\Big)\lVert v\rVert^2.
    \end{equation}
    Substituting in \eqref{eq:perturbed_energy_inequality} leads to
    \begin{equation}
        P_v V(x)  \leq a (1+\varepsilon)V(x)
        + a \Big(1+\frac1\varepsilon\Big)\lVert v\rVert^2
        +c.
    \end{equation}
\end{proof}

\begin{proof}[Proof of Propositions \ref{proposition:forced_bounded} and \ref{proposition:forced_bounded_explicit}]
    Let $\nu_t$ denote the law of $X^u_t$, so that $\nu_{t+1} = \nu_t P_{u_t}$ by definition of the perturbed chain. Fix $\varepsilon \in (0, e^\gamma - 1)$ and write
    \begin{equation}
        a_\varepsilon := a (1+\varepsilon)
        \qquad \text{and} \qquad
        \kappa_\varepsilon := a \Big( 1 + \frac{1}{\varepsilon} \Big) \in (0, \infty) ,
    \end{equation}
    where $a_\varepsilon \in (0,1)$ because $a = e^{-\gamma}$ and $\varepsilon < e^\gamma - 1$.

    For every $t \in \N$, Lemma \ref{lemma:perturbation_energy} applied with $v = u_t$ gives
    \begin{equation}
        \label{eq:perturbed_moments_onestep}
        \int V \ud \nu_{t+1}
        =
        \int P_{u_t} V \ud \nu_t
        \leq
        a_\varepsilon \int V \ud \nu_t + \kappa_\varepsilon \lVert u_t \rVert^2 + c .
    \end{equation}
    Since $\int V \ud \nu_0$ is finite by assumption, this shows by induction that $\int V \ud \nu_t$ is finite for every $t \in \N$.

    \paragraph{Iterated bound.}
    Iterating \eqref{eq:perturbed_moments_onestep} and bounding the last geometric sum by its limit yields, for every $t \in \N$,
    \begin{equation}
        \label{eq:forced_moment_iterated}
        \begin{aligned}
            \int V \ud \nu_t
             & \leq
            a_\varepsilon^{\,t} \int V \ud \nu_0
            + \kappa_\varepsilon \sum_{k=0}^{t-1} a_\varepsilon^{\,t-1-k} \lVert u_k \rVert^2
            + c \sum_{j=0}^{t-1} a_\varepsilon^{\,j}
            \\
             & \leq
            a_\varepsilon^{\,t} \int V \ud \nu_0
            + \kappa_\varepsilon \sum_{k=0}^{t-1} a_\varepsilon^{\,t-1-k} \lVert u_k \rVert^2
            + \frac{c}{1 - a_\varepsilon} .
        \end{aligned}
    \end{equation}

    Now choose $\varepsilon := (e^\gamma - 1)/2$. Then $a_\varepsilon = (1+e^{-\gamma})/2 = a'$, the value appearing in \eqref{eq:forcing_weighted_energy}, and
    \begin{equation}
        \label{eq:E_infty_explicit}
        \frac{\kappa_\varepsilon}{1 - a_\varepsilon}
        =
        \frac{2}{1 - e^{-\gamma}} \, e^{-\gamma} \Big( 1 + \frac{2}{e^\gamma - 1} \Big)
        =
        \frac{2 \, (e^\gamma + 1)}{(e^\gamma - 1)^2} ,
        \qquad
        \frac{c}{1 - a_\varepsilon}
        =
        \frac{2c}{1-a} .
    \end{equation}
    By the definition \eqref{eq:forcing_weighted_energy} of $\bar{u}_t^{\,2}$, the middle term of \eqref{eq:forced_moment_iterated} equals $\kappa_\varepsilon \, \bar{u}_t^{\,2} / (1 - a')$, so that \eqref{eq:forced_moment_iterated} is the bound \eqref{eq:forced_moment_weighted}, using $c/(1-a) = 1 + \Vert F \Vert_\infty^2/\gamma^2 + d \sigma_{\max}^2/\gamma$ as in Proposition \ref{proposition:ergodicity}.

    \paragraph{Bounded amplitude.} Suppose now that $\lVert u_t \rVert \leq U$ for every $t$. Then
    \begin{equation}
        \bar{u}_t^{\,2}
        \leq
        (1 - a') \, U^2 \sum_{k=0}^{t-1} (a')^{\,t-1-k}
        =
        \big( 1 - a'^{\,t} \big) \, U^2
        \leq U^2 ,
    \end{equation}
    and $a'^{\,t} \int V \ud \nu_0 \leq \int V \ud \nu_0$, so the right-hand side of \eqref{eq:forced_moment_weighted} is bounded uniformly in $t$, by a quantity depending on the schedule only through $U$. Since $V(x) = 1 + \Vert x \Vert^2$, this bounds $\E \big[ \Vert X_t \Vert^2 \big]$ uniformly in time, and every observable $|\varphi| \leq V$ satisfies $\big| \E [ \varphi(X_t) ] \big| \leq \int V \ud \nu_t$, which proves Proposition \ref{proposition:forced_bounded}.

    \paragraph{Stationary initialization.} For $\nu_0 = \mu$, the moment bound \eqref{eq:invariant_moment_bound} gives
    \begin{equation}
        \int V \ud \mu \leq c/(1-a) = 1 + \Vert F \Vert_\infty^2/\gamma^2 + d \sigma_{\max}^2/\gamma,
    \end{equation}
    which is finite, so the above applies. Adding this bound to the last two terms of \eqref{eq:forced_moment_weighted} gives \eqref{eq:forced_statistics_explicit}.
\end{proof}

\subsection{Proof of Proposition \ref{proposition:forced_ergodicity}}
\label{section:proofs:forced}

\forcedergodicity*

\begin{proof}
    We establish for the translated kernel $P_v$, defined by $P_v(x, \cdot) = P(x + v, \cdot)$, the three arguments that the proof of Proposition \ref{proposition:ergodicity} establishes for $P$.

    For every bounded continuous $f$, $P_v f = (Pf)(\cdot + v)$ is bounded and continuous by Lemma \ref{lemma:feller}, so $P_v$ is Feller. For every $x \in \R^d$ and every $A \in \mathcal{B}(\R^d)$ with $\mathrm{Leb}(A) > 0$, $P_v(x, A) = P(x + v, A) > 0$ by Proposition \ref{proposition:positivity}.

    Lemmas \ref{lemma:irreducible_aperiodic} and \ref{lemma:petite} therefore apply to $Q = P_v$: the perturbed chain is irreducible for the Lebesgue measure and aperiodic, and its compact sets are petite.

    Lemma \ref{lemma:perturbation_energy} with $\varepsilon := (e^\gamma - 1)/2$ gives the Foster--Lyapunov condition
    \begin{equation}
        P_v V(x) \leq a' V(x) + c'
        \qquad \text{for all } x \in \R^d ,
    \end{equation}
    with
    \begin{equation}
        a' := a (1 + \varepsilon) = \frac{1 + e^{-\gamma}}{2} \in (0, 1)
        \qquad \text{and} \qquad
        c' := a \Big( 1 + \frac{1}{\varepsilon} \Big) \Vert v \Vert^2 + c \in (0, \infty) .
    \end{equation}

    The proof of Proposition \ref{proposition:ergodicity} now applies verbatim to $P_v$, with $a$ and $c$ replaced by $a'$ and $c'$, since it uses nothing about $P$ beyond these three arguments. It yields the unique invariant probability measure $\mu_v$ and  $M_v < \infty$ and $\rho_v \in (0,1)$ such that
    \begin{equation}
        \label{eq:forced_geometric_V}
        \big\Vert P_v^t(x, \cdot) - \mu_v \big\Vert_V \leq M_v \, \rho_v^t \, V(x)
        \qquad \text{for all } x \in \R^d \text{ and } t \in \N ,
    \end{equation}
    which is the geometric ergodicity asserted by Proposition \ref{proposition:forced_ergodicity}. It also yields the moment bound
    \begin{equation}
        \label{eq:forced_moment}
        \int V \ud \mu_v \leq \frac{c'}{1 - a'} = \frac{2 (e^\gamma + 1)}{(e^\gamma - 1)^2} \Vert v \Vert^2 + \frac{2c}{1-a} ,
    \end{equation}
    where the equality follows from the definition of $c'$ and from the two identities \eqref{eq:E_infty_explicit}, whose left-hand sides are $a(1+1/\varepsilon)/(1-a')$ and $c/(1-a')$ at $\varepsilon = (e^\gamma-1)/2$, and where
    \begin{equation}
        c/(1-a) = 1 + \Vert F \Vert_\infty^2/\gamma^2 + d\sigma_{\max}^2/\gamma.
    \end{equation}

\end{proof}